\documentclass{article}

\PassOptionsToPackage{numbers, sort&compress}{natbib}
\usepackage[final]{neurips_2023}
\usepackage[thmmarks]{ntheorem}
\newtheorem{definition}{Definition}
\newtheorem{proposition}{Proposition}

\newtheorem{lemma}{Lemma}

\usepackage{setspace}

\usepackage[utf8]{inputenc} % allow utf-8 input
\usepackage[T1]{fontenc}    % use 8-bit T1 fonts
\usepackage{hyperref}       % hyperlinks
\usepackage{url}            % simple URL typesetting
\usepackage{booktabs}       % professional-quality tables
\usepackage{amsfonts}       % blackboard math symbols
\usepackage{nicefrac}       % compact symbols for 1/2, etc.
\usepackage{microtype}      % microtypography
\usepackage{xcolor}         % colors

\usepackage{amsmath}
\usepackage{tikz}
\usepackage{arydshln}
\usepackage{pifont}
\usepackage{pgfplots}
\usepackage{pgfplotstable}
\pgfplotsset{compat=newest}
\usepackage{nicematrix}
\usepackage{rotating}
\usepackage{dsfont}
\usepackage{multirow}
\usepackage{adjustbox}
\usepackage{array}
\usepackage{amssymb}

\theoremstyle{nonumberplain}
\theorembodyfont{\normalfont}
\theoremsymbol{\ensuremath\square}
\newtheorem{proof}{Proof:}

\definecolor{dodgerblue}{RGB}{30,144,255}
\definecolor{greenG}{RGB}{0,128,0}
\definecolor{color1}{RGB}{196, 164, 132} % brown
\definecolor{color2}{RGB}{30,144,255} % blue
\definecolor{color3}{RGB}{255, 16, 240} % pink

\definecolor{color_bis}{RGB}{135,206,250}
\definecolor{color_bis2}{RGB}{0,33,243}

\title{Normalization-Equivariant Neural Networks with Application to Image Denoising}

\author{%
  Sébastien Herbreteau \And  Emmanuel Moebel \And Charles Kervrann \\
  Centre Inria de l'Université de Rennes, France\\
  \texttt{\{sebastien.herbreteau, emmanuel.moebel, charles.kervrann\}@inria.fr} \\
}

\begin{document}

\maketitle

\begin{abstract}
  In many information processing systems, it may be desirable to ensure that any change of the input, whether by shifting or  scaling, results in a corresponding change in the system response.  While deep neural networks are gradually replacing all traditional automatic processing methods, they surprisingly do not guarantee such normalization-equivariance (scale + shift) property, which can be detrimental in many applications. To address this issue, we propose a methodology for adapting existing neural networks so that normalization-equivariance holds by design. Our main claim is that not only ordinary convolutional layers, but also all activation functions, including the ReLU (rectified linear unit), which are applied element-wise to the pre-activated neurons, should be completely removed from neural networks and replaced by better conditioned alternatives. To this end, we introduce affine-constrained convolutions and channel-wise sort pooling layers as surrogates and show that these two architectural modifications do preserve normalization-equivariance without loss of performance. Experimental results in image denoising show that normalization-equivariant neural networks, in addition to their better conditioning, also provide much better generalization across noise levels.
\end{abstract}

\section{Introduction}

Sometimes wrongly confused with the invariance property which designates the characteristic of a function $f$ not to be affected by a specific transformation $\mathcal{T}$ applied beforehand, the equivariance property, on the other hand, means that $f$ reacts in accordance with $\mathcal{T}$. Formally, invariance is $f \circ \mathcal{T} = f$ whereas equivariance reads $f \circ \mathcal{T} = \mathcal{T} \circ f$, where $\circ$ denotes the function composition operator. %While invariance implies an insensitivity to certain transformations, equivariance emphasizes on the other hand a certain harmony of relationships between $f$ and $\mathcal{T}$. 
Both invariance and equivariance play a crucial role in many areas of study, including physics, computer vision, signal processing and have recently been studied in various settings for deep learning-based models \cite{equivariant1, equivariant_graph2, equivariant_graph3, equivariant_graph4, equivariant_image, equivariant_image2, equivariant_image3, equivariant_image4, equivariant_pointcloud1, equivariant_pointcloud2, equivariant_pointcloud3, equivariant_translation1, equivariant_translation2}. 

In this paper, we focus on the equivariance of neural networks $f_\theta$ to a specific transformation $\mathcal{T}$, namely normalization. %Far from being solely a basic pre-processing step, normalization plays a central role in deep learning, notably for its facilitation function during training, which has been widely studied \cite{batchnorm, instancenorm, groupnorm, beyondbatchnorm}. 
Although highly desirable in many applications and in spite of its omnipresence in machine learning, current neural network architectures do not equivary to normalization. With application to image denoising, for which \textit{normalization-equivariance} is generally guaranteed for a lot of conventional methods \cite{TV, nlmeans, nlridge, OWF}, we propose a methodology for adapting existing neural networks, and in particular denoising CNNs \cite{dncnn, drunet, ffdnet, tnrd, red30}, so that \textit{normalization-equivariance} holds by design. In short, the proposed adaptation is based on two innovations:
\begin{enumerate}
    \item affine convolutions: the weights from one layer to each neuron from the next layer, \textit{i.e.} the convolution kernels in a CNN, are constrained to encode affine combinations of neurons (the sum of the weights is equal to $1$). 
    \item channel-wise sort pooling: all activation functions that apply element-wise, such as the ReLU, are substituted with higher-dimensional nonlinearities, namely two by two sorting along channels that constitutes a fast and efficient \textit{normalization-equivariant} alternative.
\end{enumerate}
Despite strong architectural constraints, we show that these simple modifications do not degrade performance and, even better, increase robustness to noise levels in image denoising both in practice and in theory.

\section{Related Work}

%Equivariant neural networks (ENN) have emerged as a promising approach to enable neural networks to better handle structured data such as images, graphs, and point clouds. 
A non-exhaustive list of application fields where equivariant neural networks were studied includes graph theory, point cloud analysis and image processing. Indeed, graph neural networks are usually expected to equivary, in the sense that a permutation of the nodes of the input graph should permute the output nodes accordingly. Several specific architectures were investigated to guarantee such a property \cite{equivariant_graph2, equivariant_graph3, equivariant_graph4}. In parallel, rotation and translation-equivariant networks for dealing with point cloud data were proposed in a recent line of research \cite{equivariant_pointcloud1, equivariant_pointcloud2, equivariant_pointcloud3}. A typical application is the ability for these networks to produce direction vectors consistent with the arbitrary orientation of the input point clouds, thus eliminating the need for data augmentation. Finally, in the domain of image processing, it may be desirable that neural networks produce outputs that equivary with regard to rotations of the input image, whether these outputs are vector fields \cite{equivariant_image}, segmentation maps \cite{equivariant_image2, equivariant_image3}, or even bounding boxes for object tracking \cite{equivariant_image4}.

%Even though \textit{equivariant} networks by design are not always the very best-performing networks when compared to their respective state-of-the-art counterparts -- the thin gap of performance resulting from the strong architectural constraints -- 
In addition to their better conditioning, 
equivariant neural networks by design  are expected to be more robust to outliers. A spectacular example has been revealed by S. Mohan \textit{et al.} \cite{homogeneous_functions} in the field of image denoising. By simply removing the additive constant (``bias'') terms in neural networks with ReLU activation functions, they showed that a much better generalization at noise levels outside the training range was ensured. Although they do not fully elucidate why biases prevent generalization, and their removal allows it, the authors establish some clues that the answer is probably linked to the  \textit{scale-equivariant} property of the resulting encoded function: rescaling the input image by a positive constant value rescales the output by the same amount.  

%In the field of image denoising, the seminal work from S. Mohan \textit{et al.} \cite{homogeneous_functions} revealed that simply removing the additive constant (``bias'') terms in neural networks with ReLU activation functions allowed for much better generalization across noise levels. Although it remains unclear why biases prevent generalization, and their removal allows it, the authors establish some clues that the answer is probably linked to the  \textit{scale-equivariant} property of the resulting encoded function: rescaling the input image by a positive constant value rescales the output by the same amount.  

%do not fully elucidate how our network achieves its remarkable generalization

%generalizes robustly across noise levels

\section{Overview of normalization-equivariance}

%In this section, we formulate the normalization equivariance property and explain to what extent this property is fundamental and desirable for image denoising. While this property is generally guaranteed in traditional image denoising methods as shown below, current deep-learning-based methods do not preserve normalization equivariance which can be detrimental in real-world applications.% such as medical or fluorescence imaging.

\subsection{Definitions and properties of three types of fundamental equivariances}

We start with formal definitions of the  different types of equivariances studied in this paper. \textcolor{black}{Please note that our definition of ``scale'' and ``shift'' may differ from the definitions given by some authors in the image processing literature.}

\begin{definition} 
A function $f : \mathbb{R}^n \mapsto \mathbb{R}^m$ is said to be:
\vspace{-0.2cm}
\begin{itemize}
\setlength\itemsep{-0.4em}
\item \textit{scale-equivariant} if $
    \: \forall x \in \mathbb{R}^n, \forall \lambda \in \mathbb{R}^+_\ast, \: f(\lambda x) = \lambda f(x)\,,
$
\item \textit{shift-equivariant} if  $\: \forall x \in \mathbb{R}^n, \forall \mu \in \mathbb{R}, \:  f(x + \mu) = f(x) + \mu\,,$
\item \textit{normalization-equivariant} if it is both 
\textit{scale-equivariant} and \textit{shift-equivariant}: 
$$
\forall x \in \mathbb{R}^n, \forall \lambda \in \mathbb{R}^+_\ast, \forall \mu \in \mathbb{R}, \: f(\lambda x + \mu) = \lambda f(x) + \mu\,, 
$$
\end{itemize}
\vspace{-0.2cm}
\noindent where addition with the scalar shift $\mu$ is applied element-wise.
\label{normalization-equivariance}
\end{definition}

 Note that the \textit{scale-equivariance} property is more often referred to as positive homogeneity in pure mathematics. Like linear maps that are completely determined by their values on a basis, the above described equivariant functions are actually entirely characterized by the values their take on specific \textcolor{black}{subsets} of $\mathbb{R}^n$, as stated by the following \textcolor{black}{lemma} (see proof in \textcolor{black}{Appendix \ref{proofs_charac}}).

\begin{lemma}[Characterizations] 
  $f : \mathbb{R}^n \mapsto \mathbb{R}^m$ is entirely determined by its values on the:
 \vspace{-0.2cm}
\begin{itemize}
\setlength\itemsep{-0.4em}
\item unit sphere $\mathcal{S}$ of $\mathbb{R}^n$ if it is \textit{scale-equivariant},
\item orthogonal complement of $\: \operatorname{Span}(\mathbf{1}_n)$, \textit{i.e.}  $\operatorname{Span}(\mathbf{1}_n)^\bot$, if it is \textit{shift-equivariant},
\item intersection $\mathcal{S} \cap \operatorname{Span}(\mathbf{1}_n)^\bot$ if it is \textit{normalization-equivariant},
\end{itemize}
\vspace{-0.2cm}
\noindent where $\mathbf{1}_n$ denotes the all-ones vector of $\mathbb{R}^n$.
\label{characterization}
\end{lemma}

%. Logically, the stricter the equivariance, the smaller the subset of determination.%Moreover, if a function has one of the three equivariance properties then $f(\vec{0}_n) = \vec{0}_m$ (assuming that $f$ is continuous for the \textit{scale-equivariant} case). 

%proof: see https://uel.unisciel.fr/mathematiques/espacevect1/espacevect1_ch04/co/apprendre_ch4_07_02.html

Finally, \textcolor{black}{Lemma} \ref{preserving} highlights three basic equivariance-preserving mathematical operations that can be used as building blocks for designing neural network architectures \textcolor{black}{(see proof in Appendix \ref{proofs_charac})}.

\begin{lemma}[Operations preserving equivariance]
Let $f$ and $g$ be two \textit{equivariant} functions of the same type (either in scale, shift or normalization). Then, subject to dimensional compatibility, all of the following functions are still \textit{equivariant}:
\vspace{-0.2cm}
\begin{itemize}
\setlength\itemsep{-0.4em}
    \item $f \circ g\:$ ($f$ composed with $g$),
    \item $x \mapsto ( f(x)^\top  \: g(x)^\top )^\top \:$ (concatenation of $f$ and $g$),
    \item $(1-t)  f +  t g\:$  for all $t\in \mathbb{R}$ (affine combination of $f$ and $g$).
\end{itemize} 
\label{preserving}
\end{lemma}

\subsection{Examples of normalization-equivariant conventional denoisers}

A (``blind'') denoiser is basically a function $f : \mathbb{R}^n \mapsto \mathbb{R}^n$ which, given a noisy image $y \in \mathbb{R}^n$, tries to map the corresponding noise-free image $x \in \mathbb{R}^n$. Since scaling up an image by a positive factor $\lambda$ or adding it up a constant shift  $\mu$ does not change its contents,  it is natural to expect \textit{scale} and \textit{shift equivariance}, \textit{i.e.} \textit{normalization equivariance}, from the denoising procedure emulated by $f$. In image denoising, a majority of methods usually assume an additive
white Gaussian noise model with variance $\sigma^2$. The corruption model then reads $y \sim \mathcal{N}(x, \sigma^2 I_n)$, where $I_n$ denotes the identity matrix of size $n$, and the noise standard deviation $\sigma > 0$ is generally passed as an additional argument to the denoiser  (``non-blind'' denoising). In this case, the augmented function $f: (y, \sigma) \in \mathbb{R}^n \times \mathbb{R}^+_\ast \mapsto \mathbb{R}^n$ is said \textit{normalization-equivariant} if:
\begin{equation} 
\forall (y, \sigma) \in \mathbb{R}^n \times \mathbb{R}^+_\ast,  \forall \lambda \in \mathbb{R}^+_\ast, \forall \mu \in \mathbb{R}, \: f(\lambda y + \mu, \lambda \sigma) = \lambda f( y ,\sigma) + \mu \,, 
\label{eq:normalization-equivariance2} 
\end{equation}
\noindent as, according to the laws of statistics, $\lambda y + \mu \sim \mathcal{N}(\lambda x + \mu, (\lambda\sigma)^2 I_n)$. In what follows, we give some well-known examples of traditional denoisers that are \textit{normalization-equivariant} (see proofs in \textcolor{black}{Appendix \ref{proofs_ne}}).

\paragraph{Noise-reduction filters:} The most rudimentary methods for image denoising are the smoothing filters, among which we can mention the averaging filter or the Gaussian filter for the linear filters and   
the median filter %(or the bilateral filter \cite{bilateral_filter})
which is nonlinear. These elementary ``blind'' denoisers all implement a \textit{normalization-equivariant} function. More generally, one can prove that a linear filter is \textit{normalization-equivariant} if and only if its coefficients add up to $1$.  In others words, \textit{normalization-equivariant} linear filters process images by affine combinations of pixels.

\paragraph{Patch-based denoising:}  %The seminal work from A. Buades \textit{et al.} \cite{nlmeans} marks the advent of a new class of denoising methods, the patch-based methods. These latter take advantage of the self-similarity: the idea that a patch rarely appears alone in a natural image and that almost perfect copies can be found in its surroundings. 
The popular N(on)-L(ocal) M(eans) algorithm \cite{nlmeans} and its variants \cite{OWF, tv-means, nlmeans_parameters} consist in computing, for each pixel, an average of its neighboring noisy pixels, weighted by the degree of similarity of \textcolor{black}{the patches to which they belong}. In other words, they \textcolor{black}{process} images by convex combinations of pixels. %, a subcategory of affine combinations where each weight is nonnegative. 
More precisely, NLM can be defined as:
\begin{equation}
f_{\operatorname{NLM}}(y, \sigma)_i = \frac{1}{W_i}\sum_{y_j \in \Omega(y_i)} e^{-\frac{\|p(y_i) - p(y_j)\|_2^2}{h^2}} y_j \quad \text{with} \quad  W_i = \sum_{y_j \in \Omega(y_i)} e^{-\frac{\|p(y_i) - p(y_j)\|_2^2}{h^2}}
\end{equation}
\noindent where $y_i$ denotes the $i^{th}$ component of vector $y$, \textcolor{black}{$\Omega(y_i)$ is the set of its neighboring pixels, $p(y_i)$ represents the vectorized patch centered at $y_i$,} and the smoothing parameter $h$ is  proportional to $\sigma$ as proposed by several authors \cite{nlmeans2, nlmeans3, nlmeans_parameters}. Defined as such, $f_{\operatorname{NLM}}$ is a \textit{normalization-equivariant} function. More recently, \textcolor{black}{N(on)-L(ocal) Ridge} \cite{nlridge}  proposes to process images by linear combinations of similar patches and achieves state-of-the-art performance in unsupervised denoising. When restricting the coefficients of the combinations to sum to $1$, that is imposing affine combination constraints, the resulting algorithm encodes a \textit{normalization-equivariant} function as well.

\paragraph{TV denoising:} Total variation (TV) denoising \cite{TV} is finally one of the most famous image denoising algorithm, appreciated for its edge-preserving properties. In its original form \cite{TV}, a TV denoiser is defined as a function $f: \mathbb{R}^n \times \mathbb{R}^+_\ast  \mapsto \mathbb{R}^n$ that solves the following equality-constrained problem:
\begin{equation}
    f_{\operatorname{TV}}(y, \sigma) = \mathop{\arg \min}\limits_{
\substack{x \in \mathbb{R}^{n}}} \: \| x \|_{\operatorname{TV}}  \quad \text{s.t.} \quad  \| y-x \|_2^2  = n\sigma^2
\end{equation}
\noindent where $\| x \|_{\operatorname{TV}} := \| \nabla x \|_2$ is the total variation of $x \in \mathbb{R}^n$. Defined as such, $f_{\operatorname{TV}}$ is a \textit{normalization-equivariant} function.

\subsection{The case of neural networks}

\input{figure_convallaria}

Deep learning hides a subtlety about normalization equivariance that deserves to be highlighted. Usually, the weights of neural networks are learned on a training set containing data all normalized to the same arbitrary interval $[a_0, b_0]$. %(\textit{e.g.} $a=0$ and $b=1$). 
This training procedure improves the performance and allows for more stable optimization of the model. At inference, unseen data are processed within the interval $[a_0, b_0]$ via a $a\hbox{-}b$ linear normalization with $a_0 \leq a < b \leq b_0$ denoted $\mathcal{T}_{a , b}$  and defined by:
\begin{equation}
 \mathcal{T}_{a , b}: y \mapsto (b-a)\frac{y - \min(y)}{\max(y) - \min(y)} + a\,.
\end{equation}
Note that this transform is actually the unique linear one with positive slope that exactly bounds the output to $[a, b]$. The data is then passed to the trained network and its response is finally returned to the original range via the inverse  operator $\mathcal{T}^{-1}_{a , b}$. This proven pipeline is actually relevant in light of the following proposition. 
\bigskip
\begin{proposition}
     $\forall \: a < b \in \mathbb{R}, \forall \: f :  \mathbb{R}^n \mapsto \mathbb{R}^m,
 \mathcal{T}^{-1}_{a , b} \circ f \circ \mathcal{T}_{a , b}$ is a \textit{normalization-equivariant} function.
 \label{case_of_nn}
\end{proposition}
While normalization-equivariance appears to be solved, a question is still remaining: how to choose the hyperparameters $a$ and $b$ for a given function $f$ ? 
Obviously, a natural choice for neural networks is to take the same parameters $a$ and $b$ as in the learning phase whatever the input image is, \textit{i.e.} $a=a_0$ and $b=b_0$, but are they really optimal? The answer to this question is generally negative. Figure \ref{photo} depicts an example of the phenomenon in image denoising, taken from a real-world application. In this example, the straightforward choice is largely sub-optimal. This suggests that there are always inherent performance leaks for deep neural networks due to the two degrees of freedom induced by the normalization (\textit{i.e.}, choice of $a$ and choice of $b$). In addition, this poor conditioning can be a source of confusion and misinterpretation in critical applications.

\subsection{Categorizing image denoisers}

Table \ref{properties_summary} summarizes the equivariance properties of several popular denoisers, either conventional \cite{TV, nlmeans, nlridge, DCT, BM3D, WNNM} or deep learning-based \cite{dncnn, drunet, nlrn, swinir, restormer}. Interestingly, if \textit{scale-equivariance} is generally guaranteed for traditional denoisers, not all of them are equivariant to shifts. In particular, the widely used algorithms DCT \cite{DCT} and BM3D \cite{BM3D} are sensitive to offsets, mainly because the hard thresholding function at their core is not \textit{shift-equivariant}. Regarding the deep-learning-based networks, only DRUNet \cite{drunet} is insensitive to scale because it is a bias-free convolutional neural network with only ReLU activation functions \cite{homogeneous_functions}. \textcolor{black}{In particular, all 
transformer models \cite{restormer, nlrn, swinir, transformer_eccv}, even bias-free, are not \textit{scale-equivariant}  due to their inherent attention-based modules.} In the next section, we show how to adapt existing neural architectures to guarantee  \textit{normalization-equivariance} without loss of performance and study the resulting class of parameterized functions $(f_\theta)$.

\addtolength{\tabcolsep}{-3.5pt} 
\begin{table}[t]
  \caption{Equivariance properties of several image denoisers (left: traditional, right: learning-based)}
  \label{properties_summary}
  \centering
  \resizebox{\columnwidth}{!}{%
  \begin{tabular}{ccccccc|ccccc}
         & TV      & NLM  & NLR & DCT & BM3D &  WNNM & DnCNN  & NLRN & SwinIR & Restormer & DRUNet \\
    \midrule
    Scale & \textcolor{greenG}{\ding{51}}
     &  \textcolor{greenG}{\ding{51}}  
     &  \textcolor{greenG}{\ding{51}} &  \textcolor{greenG}{\ding{51}}  & \textcolor{greenG}{\ding{51}} & \textcolor{greenG}{\ding{51}}
     &  \textcolor{red}{\ding{55}}  & \textcolor{red}{\ding{55}} &  \textcolor{red}{\ding{55}} &  \textcolor{red}{\ding{55}} & \textcolor{greenG}{\ding{51}} \\
    Shift &  \textcolor{greenG}{\ding{51}}
     & \textcolor{greenG}{\ding{51}} & \textcolor{greenG}{\ding{51}}
     &  \textcolor{red}{\ding{55}}  & \textcolor{red}{\ding{55}} & \textcolor{red}{\ding{55}} & \textcolor{red}{\ding{55}}
       & \textcolor{red}{\ding{55}} &  \textcolor{red}{\ding{55}} & \textcolor{red}{\ding{55}} & \textcolor{red}{\ding{55}} \\
    \bottomrule
  \end{tabular}}
\end{table}
\addtolength{\tabcolsep}{3.5pt} 

\section{Design of Normalization-Equivariant Networks}

%In this section, we explain how to adapt existing neural architectures to guarantee \textit{normalization-equivariance} 

\subsection{Affine convolutions}

To justify the introduction of a new type of convolutional layers, let us study one of the most basic neural network, namely the linear (parameterized) function $f_\Theta : x \in \mathbb{R}^{n} \mapsto \Theta x$, where parameters $\Theta$ are a matrix of $\mathbb{R}^{m \times n}$. Indeed, $f_\Theta$ can be interpreted as a dense neural network with no bias, no hidden layer and no activation function. Obviously, $f_\Theta$ is always $\textit{scale-equivariant}$, whatever the weights $\Theta$. As for the $\textit{shift-equivariance}$, a simple calculation shows that:
\begin{equation}
    x\mapsto \Theta x \text{ is } \textit{shift-equivariant} \: \Leftrightarrow \: \forall x \in \mathbb{R}^n, \forall \mu \in \mathbb{R}, \Theta(x + \mu  \mathbf{1}_n) = \Theta x + \mu \mathbf{1}_m \: \Leftrightarrow \: \Theta \mathbf{1}_n = \mathbf{1}_m\,.
    \label{eq:affine_demo}
\end{equation}

Therefore,  $f_\Theta$ is \textit{normalization-equivariant} if and only if each row of matrix $\Theta$ sums to $1$. In other words, for the \textit{normalization-equivariance} to hold, the rows of $\Theta$ must encode weights of affine combinations. \color{black} Transposing the demonstration to any convolutional neural network follows from the observation that a convolution from an input layer of size $H \times W \times C$ to an output layer of size $H' \times W' \times C'$ can always be represented with a dense connection by vectorizing the input and output layers. The elements of the $C'$ convolutional kernels of size $k \times k \times C$ each are then stored separately along the rows of the (sparse) transition matrix $\Theta$ of size $(H' \times W' \times C') \times (H \times W \times C)$. Therefore, a convolutional layer preserves the \textit{normalization-equivariance} if and only if the weights of the each convolutional
kernel  sums to $1$.
\color{black} In the following, we call such convolutional layers ``affine convolutions''.

\color{black}
In order to guarantee the affine constraint on each convolutional kernel throughout the training phase, one possibility is to ``telescope'' the circular shifted version of an unconstrained kernel to itself (this way, the sum of the resulting trainable coefficients cancels out) and then add the inverse of the kernel size element-wise as a non-trainable offset. Despite this over-parameterized form (involving an extra degree of freedom), we found this solution to be easier to use in practice. Moreover, it ensures that all coefficients of the affine kernels follow the same law at initialization. 
\color{black}

Since \textit{normalization-equivariance} is preserved through function composition, concatenation and affine combination (see \textcolor{black}{Lemma} \ref{preserving}), a (linear) convolutional neural network composed of only affine convolutions with no bias and possibly skip or \textit{affine} residual connections (trainable affine combination of two layers), is guaranteed to be \textit{normalization-equivariant}, provided that padding is performed with existing features (reflect, replicate or circular padding for example). Obviously, in their current state, these neural networks are of little interest, as linear functions do not encode best-performing functions for many applications, image denoising being no exception. Nevertheless, based on such networks, we show in the next subsection how to introduce nonlinearities without breaking the \textit{normalization-equivariance}.

%This observation legitimates the introduction of what we call affine convolutions.

\subsection{Channel-wise sort pooling as a normalization-equivariant alternative to ReLU}

The first idea that comes to mind is to apply a nonlinear activation function $\varphi : \mathbb{R} \mapsto \mathbb{R}$  preserving \textit{normalization-equivariance} after each affine convolution. In other words, we look for a nonlinear solution $\varphi$ of the characteristic functional equation of \textit{normalization-equivariant} functions (see Def. \ref{normalization-equivariance}) for $n=1$. \color{black} Unfortunately, according to Prop. \ref{set_prop} (see proof in Appendix \ref{proofs_charac} which is based on Lemma \ref{characterization}), the unique solution is the identity function which is linear. Therefore, activation functions that apply element-wise are to be excluded. 

\newpage
\begin{proposition}
Let $\operatorname{NE}(n)$ be the set of \textit{normalization-equivariant} functions from 
$\mathbb{R}^n$ to $\mathbb{R}^n$. 

$\operatorname{NE}(1) = \{ x \mapsto x \}\,$ and

$\operatorname{NE}(2) =\left \{  {\left. (x_1, x_2)  \mapsto A \begin{pmatrix} x_1 \\ x_2 \end{pmatrix}  \mbox{ if } x_1 \leq x_2 \mbox{ else }  B \begin{pmatrix}    x_1 \\ x_2 \end{pmatrix}     
 \,\right|\, A, B \in \mathbb{R}^{2 \times 2}} \;\text{s.t.}\; A \mathbf{1}_2 = B \mathbf{1}_2 = \mathbf{1}_2 \right \}.$
\label{set_prop}
\end{proposition}

To find interesting nonlinear functions, one needs to examine multi-dimensional activation functions, \textit{i.e.} ones of the form $\varphi : \mathbb{R}^n \mapsto \mathbb{R}^m$ with $n\geq 2$. In order to preserve the dimensions of the neural layers and to limit the computational costs, we focus on the case $n=m=2$, meaning that $\varphi$ processes pre-activated neurons by pairs. According to Prop. \ref{set_prop}, the \textit{normalization-equivariant} functions from $\mathbb{R}^2$ to $\mathbb{R}^2$ are parameterized by two matrices $A,B \in \mathbb{R}^{2 \times 2}$ such that $A \mathbf{1}_2 = B \mathbf{1}_2 = \mathbf{1}_2$ and apply a different (affine-constrained) linear mapping depending on whether or not the input is in ascending order. As long as $A\neq B$, the resulting function is nonlinear and makes it \textit{de facto} a candidate to replace the conventional one-dimensional activation functions such as the popular ReLU (rectified linear unit) function. Interestingly, when arbitrarily choosing  $A$ and $B$ to be the permutation matrices of $\mathbb{R}^{2 \times 2}$, the resulting \textit{normalization-equivariant} function simply reads:
\begin{equation}
    \varphi: (x_1,x_2) \in \mathbb{R}^2 \mapsto \begin{pmatrix} \min(x_1,x_2) \\ \max(x_1,x_2) \end{pmatrix}\,,
    \label{eq:sortpool_varphi}
\end{equation}
which is nothing else than the sorting function in $\mathbb{R}^2$. Clearly, it is among the simplest \textit{normalization-equivariant} nonlinear function from $\mathbb{R}^2$ to $\mathbb{R}^2$ and it is the one we consider as surrogate for the one-dimensional activation functions (choosing other functions, that is considering other choices for $A$ and $B$, does not bring improvements in terms of performance in our experiments). More generally, it is easy to show that all the sorting functions of $\mathbb{R}^n$ are \textit{normalization-equivariant} and are nonlinear as soon as $n\geq2$. 
\textcolor{black}{Note that such sorting operators have been promoted by \cite{oplu, lipschitz} in totally different contexts for their norm-preserving properties of the backpropagated gradients.}

\color{black}

% By arbitrarily setting $y_{-u} = y_{u} = \begin{pmatrix} \alpha // \beta \end{pmatrix}$ where $\alpha$ and $\beta$ are trainable parameters, we have:
%\begin{equation} \varphi: (x_1, x_2) \in \mathbb{R}^2 \mapsto | x_2 - x_1 |\begin{pmatrix} \alpha / \sqrt{2} // \beta / \sqrt{2} \end{pmatrix} + \frac{x_1+x_2}{2} \mathbf{1}_2 \,,\end{equation}

 %The good news is that these functions are nonlinear as soon as $n\geq2$. Therefore, they are candidates to replace the conventional activation functions such as the popular ReLU (rectified linear unit) function.

%Indeed, the sorting function (\ref{sortpool_varphi}) is a normalization-equivariant function by concatenation (see Prop. \ref{preserving}) of two normalization-equivariant functions from $\mathbb{R}^2$ to $\mathbb{R}$, namely the functions min and max. Moreover, it does verify $\varphi(-u) = \varphi(u) = u$ and this is the only one according to Prop. \ref{characterization}. Note that another choice of arbitrary setting would have entirely identified a different normalization-equivariant function of $\mathbb{R}^2$, possibly more complicated and especially less trivial to derive.

\begin{figure}[t]
    \centering
\includegraphics[width=\columnwidth]{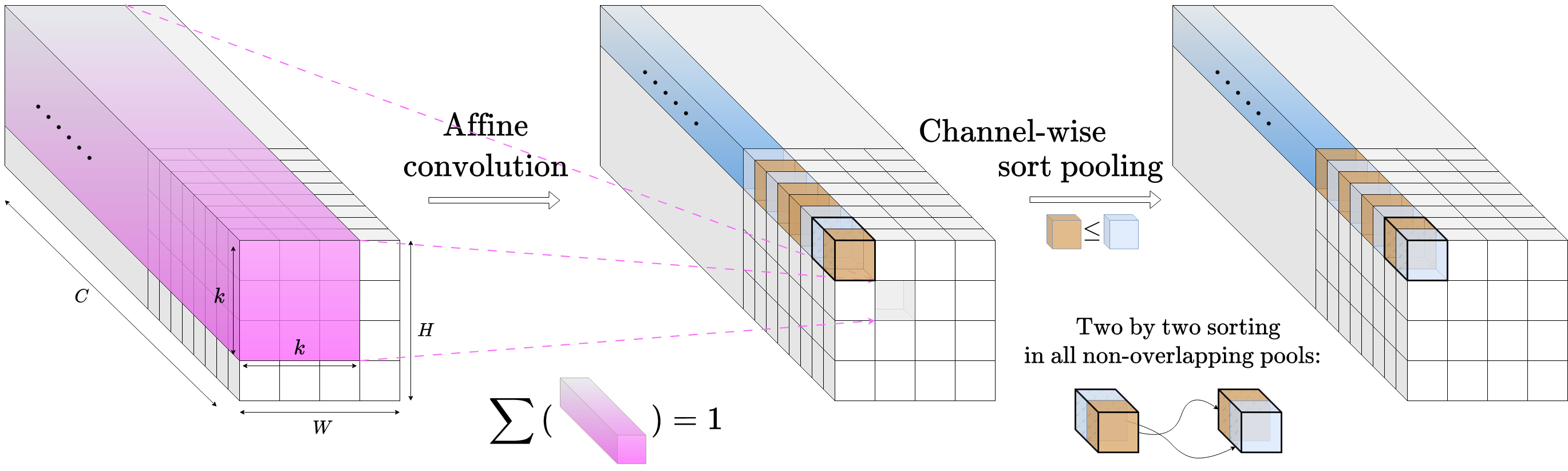}
    \caption{Illustration of the proposed alternative for replacing the traditional scheme ``convolution + element-wise activation function'' in convolutional neural networks:
    affine convolutions supersede ordinary ones by restricting the coefficients of each kernel to sum to one and the proposed sort pooling patterns introduce nonlinearities by sorting two by two the pre-activated neurons along the channels.}
    \label{sortpooling}
\end{figure}

Since the sorting function \eqref{eq:sortpool_varphi} is to be applied on non-overlapping pairs of neurons, the partitioning of layers needs to be determined. In order not to mix unrelated neurons, we propose to apply this two-dimensional activation function channel-wisely across layers and call this operation ``sort pooling'' in reference to the max pooling operation, widely used for downsampling, and from which it can be effectively implemented. Figure \ref{sortpooling} illustrates the sequence of the two proposed innovations, namely affine convolution followed by channel-wise sort pooling, to replace the traditional scheme ``conv+ReLU'', while guaranteeing \textit{normalization-equivariance}.

\subsection{Encoding adaptive affine filters}

Based on \textcolor{black}{Lemma} \ref{preserving}, we can formulate the following proposition which tells more about the class of parameterized functions $(f_\theta)$ encoded by the proposed networks. 
\newpage

\begin{proposition}
Let $f_\theta^{\text{NE}} : \mathbb{R}^n \mapsto \mathbb{R}^m$ be a CNN composed of only:
\vspace{-0.2cm}
\begin{itemize}
\setlength\itemsep{-0.4em}
    \item affine convolution kernels with no bias and where padding is made of existing features,
    \item sort pooling nonlinearities,
    \item possibly skip or affine residual connections\textcolor{black}{, and max or average pooling layers}.
\end{itemize}
\vspace{-0.2cm}
\noindent Then, $f_\theta^{\text{NE}}$ is a normalization-equivariant continuous piecewise-linear function with finitely many pieces. Moreover, on each piece represented by the vector $y_{r}$,  \vspace{-0.2cm}
    \begin{center}
        $f_\theta^{\text{NE}}(y) = A_\theta^{y_{r}} y, \: \text{ with } A_\theta^{y_{r}} \in \mathbb{R}^{m \times n} \text{ such that } A_\theta^{y_{r}} \mathbf{1}_n = \mathbf{1}_m\,.$
    \end{center} 
\label{theorem}
\end{proposition}
\vspace{-0.2cm}
In Prop. \ref{theorem}, the subscripts on $A_\theta^{y_{r}}$ serve as a reminder that this matrix depends on the sort pooling activation patterns, which in turn depend on both the input vector $y$ and the weights $\theta$. As already revealed for bias-free networks with ReLU \cite{homogeneous_functions}, $A_\theta^{y_{r}}$ is the Jacobian matrix of $f_\theta^{\text{NE}}$ taken at any point $y$ in the interior of the piece represented by vector $y_r$. Moreover, as $A_\theta^{y_{r}} \mathbf{1}_n = \mathbf{1}_m$, the output vector of such networks are locally made of fixed affine combinations of the entries of the input vector. And since a CNN has a limited receptive field centered on each pixel, $f_\theta^{\text{NE}}$ can be thought of as an adaptive filter that produces an estimate of each pixel through a custom affine combination of pixels. By examining these filters in the case of image denoising (see Fig. \ref{photo2}), it becomes apparent that they vary in their characteristics and are intricately linked to the contents of the underlying images. Indeed, these filters are specifically designed to cater to the specific local features of the noisy image: averaging is done over uniform areas without affecting the sharpness of edges. Note that this behavior has already been extensively studied by \cite{homogeneous_functions} for unconstrained filters.

The total number of fixed adaptive affine filters depends on the weights $\theta$ of the network $f_\theta^{\text{NE}}$ and is bounded by $2^{S}$ where $S$ represents the total number of sort pooling patterns traversed to get from the receptive filed to its final pixel \textcolor{black}{(assuming no max pooling layers)}. Obviously, this upper bound grows exponentially with $S$, suggesting that a limited number of sort pooling operations may generate an extremely large number of filters. Interestingly, if ReLU activation functions where used instead, the upper bound would reach $2^{2S}$.

\input{figure_house_short_short}

\section{Experimental results}

We demonstrate the effectiveness and versatility of the proposed methodology in the case of image denoising. To this end, we modify two well-established neural network architectures for image denoising, chosen for both their simplicity and efficiency, namely DRUNet \cite{drunet}: a state-of-the-art U-Net with residual connections \cite{resnet}; and FDnCNN, the unpublished flexible variant of the popular DnCNN \cite{dncnn}: a simple feedforward CNN that chains ``conv+ReLU'' layers with no downsampling, no residual connections and no batch normalization during training \cite{batchnorm}, and with a tunable noise level map as additional input \cite{ffdnet}. We show that adapting these networks to become \textit{normalization-equivariant} does not adversely affect performance and, better yet, increases their generalization capabilities. For each scenario, we train three variants of the original Gaussian denoising network for grayscale images: \textit{ordinary} (original network with additive bias), \textit{scale-equivariant} (bias-free variation with ReLU  \cite{homogeneous_functions}) and our \textit{normalization-equivariant} architecture (see Fig. \ref{sortpooling}). Details about training and implementations can be found in  \textcolor{black}{Appendix \ref{appendix_A} and Appendix \ref{appendix_B}}. Unless otherwise noted, all results presented in this paper are obtained with DRUNet \cite{drunet}; similar outcomes can be achieved with FDnCNN \cite{dncnn} architecture (see \textcolor{black}{Appendix \ref{appendix_C}}).

Finally, note that both DRUNet \cite{drunet} and FDnCNN \cite{dncnn} can be trained as ``blind'' but also as ``non-blind'' denoisers and thus achieve increased performance, by passing an additional noisemap as input. In the case of additive white Gaussian noise of variance $\sigma^2$, the noisemap is constant equal to $\sigma \mathbf{1}_n$ and the resulting parameterized functions can then be  put mathematically under the form $f_\theta: (y, \sigma) \in \mathbb{R}^n \times \mathbb{R}^+_\ast \mapsto \mathbb{R}^n$. In order to integrate this feature to \textit{normalization-equivariant} networks as well, a slight modification of the first affine convolutional layer must be made. Indeed, by adapting the proof \eqref{eq:affine_demo} to the case \eqref{eq:normalization-equivariance2}, we can show that the first convolutional layer must be affine with respect to the input image $y$ only -- the coefficients of the kernels acting on the image pixels add up to $1$ -- while the other coefficients of the kernels need not be constrained.

%However, for \textit{normalization-equivariance} to remain valid in this case, the first convolutional layer must be affine with respect to the input image $y$ only -- the coefficients of the kernels acting on the image pixels add up to $1$ -- while the other coefficients of the kernels need not be constrained, as shown by the adaptation of the proof (\ref{affine_demo}) to the case (\ref{normalization-equivariance2}).  

\subsection{The proposed architectural modifications do not degrade performance}

 The performance, assessed in terms of PSNR values, of our \textit{normalization-equivariant} alternative (see Fig. \ref{sortpooling}) and of its \textit{scale-equivariant} and \textit{ordinary} counterparts  is compared in Table \ref{resultsPSNR} for ``non-blind'' architectures on two popular datasets \cite{berkeley}. We can notice that the performance gap between two different variants is less than 0.05 dB at most for all noise levels, which is not significant. This result suggests that the class of parameterized functions $(f_\theta)$ currently used in image denoising can drastically be reduced at no cost. 
Moreover, it shows that it is possible to dispense with activation functions, such as the popular ReLU:
nonlinearities can simply be brought by sort pooling patterns. In terms of subjective visual evaluation, we can draw the same conclusion since images produced by two architectural variants inside the training range are hardly distinguishable (see Fig. \ref{photo2} at $\sigma=25$).

\subsection{Increased robustness across noise levels}
\label{subsection_noise}

\input{figure_generalization}
\begin{table*}[t]
\centering
\caption{The PSNR (dB) results of ``non-blind'' deep learning-based methods applied to popular grayscale datasets corrupted by synthetic white Gaussian noise with $\sigma=15$, $25$ and $50$.}

\resizebox{0.8\columnwidth}{!}{%
  \begin{NiceTabular}{c@{\hspace{0.1cm}}  c@{\hspace{0.5cm}}
c@{\hspace{0.1cm}} c@{\hspace{0.1cm}} c@{\hspace{0.1cm}} c@{\hspace{0.1cm}}  c@{\hspace{0.7cm}}
c@{\hspace{0.1cm}} c@{\hspace{0.1cm}} c@{\hspace{0.1cm}} c@{\hspace{0.1cm}}  
 c@{\hspace{0.0cm}}  c}
  \hline
\multicolumn{2}{c}{Dataset}  & \multicolumn{5}{c}{Set12} & \multicolumn{5}{c}{BSD68} &  \\\hline\hline\noalign{\vskip 0.1cm}

   \multicolumn{2}{c}{ Noise level $\sigma$} &  15 &/& 25  &/& 50 &  15 &/& 25  &/& 50  &  \\[0.1cm]
   \hline\noalign{\vskip 0.1cm}

 \multirow{3}{*}{\small DRUNet \cite{drunet}} \quad \quad  & \textit{ordinary}   &   33.23 &/& 30.92  &/& 27.87 &  31.89 &/& 29.44  &/& 26.54 &\\
 & \textit{scale-equiv}  &   33.25 &/& 30.94  &/& 27.90 &   31.91 &/& 29.48  &/& 26.59 &\\
 & \textbf{\textit{norm-equiv}}  &   33.20 &/& 30.90  &/& 27.85 &   31.88 &/& 29.45  &/& 26.55   &\\

\cdashline{1-12}\noalign{\vskip 0.1cm}
%\hline\noalign{\vskip 0.1cm}

   \multirow{3}{*}{\small FDnCNN \cite{dncnn}} \quad \quad  & \textit{ordinary}  &   32.87 &/& 30.49  &/& 27.28 &   31.69 &/& 29.22  &/& 26.27 &\\
 & \textit{scale-equiv}  &   32.85 &/& 30.49  &/& 27.29 &   31.67 &/&  29.20 &/& 26.25 &\\
 & \textbf{\textit{norm-equiv}}  &    32.85 &/& 30.50  &/& 27.27 &   31.69 &/& 29.22  &/& 26.25 & \\[0.1cm] \hline
\end{NiceTabular}%
}

%\small $\ast$: after variance-stabilizing transformation.
\label{resultsPSNR}
\end{table*}

S. Mohan \textit{et al.} \cite{homogeneous_functions} revealed that bias-free neural networks with ReLU, which are \textit{scale-equivariant}, could much better generalize when evaluated at new noise levels beyond their training range, than their counterparts with bias that systematically overfit. Even if they do not fully elucidate how such networks achieve this remarkable generalization, they suggest that \textit{scale-equivariance} certainly plays a major role. What about \textit{normalization-equivariance} then? We have compared the robustness faculties of the three variants of networks when trained at a fixed noise level $\sigma$ for Gaussian noise. 
Figure \ref{noise_generalization}  summarizes the explicit results obtained: \textit{normalization-equivariance} pushes generalization capabilities of neural networks one step further. While performance is identical to their \textit{scale-equivariant} counterparts when evaluated at higher noise levels, the \textit{normalization-equivariant} networks are, however, much more robust at lower noise levels. This phenomenon is also illustrated in Fig. \ref{photo2}.

\input{figure_monarch}

\paragraph{Demystifying robustness} 
%Since CNNs operate within a limited area known as the receptive field, it is legitimate to confine the following argument solely to image patches.
Let $x$ be a clean patch of size $n$, representative of the training set on which a CNN $f_\theta$ was optimized to denoise its noisy realizations $y = x + \varepsilon$ with $\varepsilon \sim \mathcal{N}(0, \sigma^2 I_n)$ (denoising at a fixed noise level $\sigma$ exclusively). Formally, we note $x \in \mathcal{D} \subset \mathbb{R}^n$, where $\mathcal{D}$ is the
% hypothetical (infinite) 
space of representative clean patches of size $n$ on which $f_\theta$ was trained. We are interested in the output of $f_\theta$ when it is evaluated at $x+\lambda \varepsilon$ (denoising at noise level $\lambda \sigma$) with $\lambda > 0$. % supposed unknown to the user. 
Assuming that $f_\theta$ encodes a \textit{normalization-equivariant} function, we have:
\begin{equation}
    \forall \lambda \in \mathbb{R}^+_\ast, \forall \mu \in \mathbb{R}, \: f_\theta(x+\lambda \varepsilon) = \lambda    f_\theta((x - \mu)/\lambda  +\varepsilon) + \mu \,.
\end{equation}
The above equality shows how such networks can deal with noise levels $\lambda \sigma$ different from $\sigma$: \textit{normalization-equivariance} simply brings the problem back to the denoising of an implicitly renormalized image patch with fixed noise level $\sigma$. %Indeed, denoising an image $x$ corrupted by Gaussian noise of standard deviation $\lambda\sigma$ amounts to denoising  $(x - \mu) /\lambda$ with noise level $\sigma$. 
Note that this artificial change of noise level does not make this problem any easier to solve as the signal-to-noise ratio is preserved by normalization. Obviously, the denoising result of $x+\lambda\varepsilon$ will be all the more accurate as $(x - \mu) /\lambda$ is a representative patch of the training set. In other words, if $(x - \mu) /\lambda$ can still be considered to be in $\mathcal{D}$, then $f_\theta$ should output a consistent denoised image patch. For a majority of methods \cite{dncnn, drunet, ffdnet}, training is performed within the interval $[0, 1]$ and therefore $x / \lambda$ still belongs generally to $\mathcal{D}$ for $1 < \lambda < 10$ (contraction), but this is much less true for $\lambda < 1$ (stretching) for the reason that it may exceed the bounds of the interval $[0, 1]$. This explains why \textit{scale-equivariant} functions do not generalize well to noise levels lower than their training one. In contrast, \textit{normalization-equivariant} functions can benefit from the implicit extra adjustment parameter $\mu$. Indeed, there exists some cases where the stretched patch $x/\lambda$ is not in $\mathcal{D}$ but $(x - \mu) /\lambda$ is (see Fig. \ref{monarch}b). This is why \textit{normalization-equivariant} networks are more able to generalize at low noise levels. Note that, based on this argument, \textit{ordinary} neural networks trained at a fixed noise level $\sigma$ can also be used to denoise images at noise level $\lambda \sigma$, provided that a correct normalization is done beforehand \cite{normalization_morel}. However, this time the normalization is explicit: the exact scale factor $\lambda$, and possibly the shift $\mu$, must be known (see Fig. \ref{monarch}a).

\color{black}
It turns out that this theoretical argument is valid for a wide range of noise types, not only Gaussian noise. Indeed, the same argument holds for any additive noise $\varepsilon$ that possesses the scaling property: $\lambda \varepsilon$ belongs to the same family of probability distributions as $\varepsilon$ (e.g., Gaussian, uniform, Laplace or even Rayleigh noise which is not zero-mean). By the way, the authors of \cite{homogeneous_functions} had already verified the noise generalization capabilities of \textit{scale-equivariant} networks for uniform noise in addition to Gaussian noise, without fully elucidating why it works. In Appendix \ref{appendix_C}, we checked experimentally that ``blind'' \textit{normalization-equivariant} networks trained on additive uniform, Laplace or Rayleigh noise at a single noise level are much more robust at unseen noise levels than their \textit{scale-equivariant} and \textit{ordinary} counterparts.
\color{black}

 %Moreover, we also observed the robustness capabilities of \textit{normalization-equivariant} networks when testing on a type of noise that is more difficult to characterize, namely JPEG noise, or JPEG artifacts (see Supplementary Material)

\section{Conclusion and perspectives}

In this work, we presented an original approach to adapt the architecture of existing neural networks so that they become \textit{normalization-equivariant}, a property highly desirable and expected in many applications such that image denoising. We argue that the classical pattern  ``conv+ReLU'' can be favorably replaced by the two proposed innovations: affine convolutions that ensure that all coefficients of the convolutional kernels sum to one; and channel-wise sort pooling nonlinearities as a substitute for all activation functions that apply element-wise, including ReLU or sigmoid functions. Despite these two important architectural changes, we show that the performance of these alternative networks is not affected in any way. On the contrary, thanks to their better-conditioning, they benefit, in the context of image denoising, from an increased interpretability and especially robustness to variable noise levels both in practice and in theory.  

%\textcolor{orange}{More generally, the proposed channel-wise sort pooling nonlinearities may potentially change the way we  commonly understand neural networks: the usual paradigm that neurons are either active (``fired'') or inactive, is indeed somewhat shaken. With sort pooling nonlinearities, neurons are no longer static but they ``wiggle and mingle'' according to the received signal. We believe that this discovery may help building new neural architectures, potentially with stronger theoretical guarantees, and more broadly, may also open the doors for novel perspectives in deep learning.}

\color{black}

\section*{Limitations}

We would like to mention that the proposed architectural modifications for enforcing \textit{normalization-equivariance} require a longer training for achieving comparable performance with its original counterparts (see Appendix \ref{appendix_B}), and may be incompatible with some specific network layers such as batch-norm \cite{batchnorm} or attention-based modules \cite{restormer, nlrn, swinir, transformer_eccv}. Moreover, our method has shown its potential mainly to image denoising as it stands, even though in principle \textit{normalization-equivariance} may be applicable and helpful in other tasks as well (see preliminary results about image classification in Appendix \ref{appendix_C}). Discovering similar advantages of \textit{normalization-equivariance} in other computer vision tasks, possibly related to outlier robustness, is an interesting avenue of research for future work.

\color{black}

\begin{ack}
This work was supported by Bpifrance agency (funding) through the LiChIE contract. Computations  were performed on the Inria Rennes computing grid facilities partly funded by France-BioImaging infrastructure (French National Research Agency - ANR-10-INBS-04-07, ``Investments for the future'').
We would like to thank R. Fraisse (Airbus) for fruitful  discussions. 
\end{ack}

% Bibliography
%\input{bib.tex}
%\newpage
\bibliography{bib.bib}

\newpage
\appendix
\section{Description of the denoising architectures and implementation}
\label{appendix_A}

\subsection{Description of models}

\paragraph{DRUNet:}

DRUNet \cite{drunet} is a U-Net architecture, and as such has an encoder-decoder type pathway, with residual connections \cite{resnet}. Spatial downsampling is performed using $2 \times 2$ convolutions with stride $2$, while spatial upsampling leverages $2 \times 2$ transposed convolutions with stride $2$ (which is equivalent to a $1 \times 1$ sub-pixel convolution \cite{pixelshuffle}). The number of channels in each layer from the first scale to the fourth scale are $64$, $128$, $256$ and $512$, respectively. Each scale is composed of 4 successive residual blocks ``$3\times3$ conv + ReLU + $3\times3$ conv''.

\paragraph{FDnCNN:}

FDnCNN \cite{dncnn} is the unpublished flexible variant of the popular DnCNN \cite{dncnn}. It consists of 20 successive $3\times3$ convolutional layers with $64$ channels each and ReLU nonlinearities. As opposed to DnCNN, FDnCNN does not use neither batch normalization \cite{batchnorm} for training, nor residual connections \cite{resnet} and can handle an optional noisemap (concatenated with the input noisy image). Note that this architecture does not use downsampling or upsampling. Finally, the authors \cite{dncnn} recommend to train it by minimizing the $\ell_1$ loss instead of the mean squared error (MSE).

\subsection{Description of variants}

\paragraph{Ordinary:} The \textit{ordinary} variant is built by appending additive constant (``bias'') terms after each convolution of the original architecture. Note that the original FDnCNN \cite{dncnn} model is already in the \textit{ordinary} mode.

\paragraph{Scale-equivariant:} Since both models (DRUNet and FDnCNN) use only ReLU activation functions, removing all additive constant (``bias'') terms is sufficient to ensure \textit{scale-equivariance} \cite{homogeneous_functions}. Note that the original DRUNet \cite{drunet} model is already in the \textit{scale-equivariant} mode.

\paragraph{Normalization-equivariant:} All convolutions are replaced by the proposed affine-constrained convolutions without ``bias'' and with reflect padding, and the proposed channel-wise sort pooling patterns supersede ReLU nonlinearities. Moreover, classical residual connections are replaced by \textit{affine} residual connections (the sum of two layers $l_1$ and $l_2$ is replaced by their affine combination $(1-t)  l_1 + t  l_2$ where $t$ is a trainable scalar parameter).

\subsection{Practical implementation of normalization-equivariant networks}

The channel-wise sort pooling operations can be efficiently implemented by concatenating the sub-layer obtained with channel-wise one-dimensional max pooling with kernel size $2$ and its counterpart obtained with min pooling. Note that intertwining these two sub-layers to comply with the original definition is not necessary in practice (although performed anyway in our implementation), since the order of the channels in a CNN is arbitrary. 

Regarding the implementation of affine convolutions for training, each unconstrained kernel can be in practice ``telescoped''  with its circular shifted version (this way, the sum of the resulting trainable coefficients cancels out) and then the inverse of the kernel size is added element-wise as a non-trainable offset. Despite this over-parameterized form (involving an extra degree of freedom), we found this solution to be more easy to use in practice. Moreover, it ensures that all coefficients of the affine kernels follow the same law at initialization. Another possibility is to set an arbitrary coefficient of the kernel (the last one for instance) equal to one minus the sum of all the other coefficients. Note that the solution consisting in dividing each kernel coefficient by the sum of all the other coefficients does not work because it generates numerical instabilities as the divisor may be zero, or close to zero.

\textcolor{black}{All our implementations are written in Python and are based on the PyTorch library \cite{pytorch}. The code and pre-trained models can be downloaded
here: \href{https://github.com/sherbret/normalization_equivariant_nn/}{\nolinkurl{https://github.com/sherbret/normalization_equivariant_nn/}}.}

\section{Description of datasets and training details}
\label{appendix_B}

\begin{table*}[t]
\centering
\caption{Training parameters. * indicates that it is divided by half every $100,000$ iterations.}

\resizebox{1.00\columnwidth}{!}{%
  \begin{NiceTabular}{c@{\hspace{0.1cm}}  c@{\hspace{0.5cm}} c@{\hspace{0.7cm}} c@{\hspace{0.7cm}} c@{\hspace{0.7cm}} c@{\hspace{0.7cm}}
 c@{\hspace{0.0cm}}}
  \hline
\multicolumn{2}{c}{Model}  & \begin{tabular}{c}
Batch \\ size
\end{tabular}  & \begin{tabular}{c}
Patch \\ size
\end{tabular}  & \begin{tabular}{c}
Loss \\ function
\end{tabular} &  \begin{tabular}{c}
Learning \\ rate
\end{tabular}  & \begin{tabular}{c}
Number of \\ iterations
\end{tabular} \\
   \hline\noalign{\vskip 0.1cm}

 \multirow{3}{*}{\small DRUNet \cite{drunet}} \quad \quad  & \textit{ordinary}   &  $16$   & $128 \times 128$  & $\ell_1$ &  $1e\hbox{-}4$* & $800,000$  \\
 & \textit{scale-equiv} &  $16$   & $128 \times 128$  & $\ell_1$ & $1e\hbox{-}4$* & $800,000$  \\
 & \textit{norm-equiv}  &  $16$   & $128 \times 128$  & \small MSE & $1e\hbox{-}4$ & $1,800,000$  \\

\cdashline{1-7}\noalign{\vskip 0.1cm}
%\hline\noalign{\vskip 0.1cm}

   \multirow{3}{*}{\small FDnCNN \cite{dncnn}} \quad \quad  & \textit{ordinary}  &  $128$   & $70 \times 70$  & $\ell_1$ & $1e\hbox{-}4$ & $500,000$  \\
 & \textit{scale-equiv} &  $128$   & $70 \times 70$  & $\ell_1$ & $1e\hbox{-}4$ & $500,000$  \\
 & \textit{norm-equiv} &  $128$   & $70 \times 70$  & \small MSE & $1e\hbox{-}4$ & $900,000$  \\[0.1cm] \hline
\end{NiceTabular}%
}

\label{training_params}
\end{table*}

We use the same large training set as in \cite{drunet} for all the models and all the experiments, composed of $8,694$ images, including $400$ images from the Berkeley Segmentation Dataset BSD400 \cite{berkeley}, $4,744$ images from the Waterloo Exploration Database \cite{dset_waterloo}, $900$ images from the DIV2K dataset \cite{dset_div2k}, and $2,750$ images from the Flickr2K dataset \cite{dset_flickr2k}. This training set is augmented via random vertical and horizontal flips and random $90^{\circ}$ rotations. The dataset BSD32 \cite{berkeley}, composed of the $32$ images, is used as validation set to control training and select the best model at the end. Finally, the two datasets Set12 and BSD68 \cite{berkeley}, strictly disjoint from the training and validation sets, are used for testing.

All the models $f_\theta$ are optimized by minimizing the average reconstruction error between the denoised images $\hat{x} = f_\theta(x+\varepsilon)$, where $\varepsilon \sim \mathcal{N}(0, \sigma^2 I_n)$, and ground-truths $x$ with Adam algorithm \cite{adam}. For ``non-blind'' models, the noise level  $\sigma$ is randomly chosen from $[1, 50]$ during
training. The training parameters, specific to each model and its variants, are guided by the instructions of the original papers \cite{drunet, dncnn}, to the extent possible, and are summarized in Table \ref{training_params}. Note that each training iteration consists in a gradient pass on a batch composed of patches randomly cropped from training images.
\textit{Normalization-equivariant} variants need a longer training and always use a constant learning rate (speed improvements are however certainly possible by adapting the learning rate throughout optimization, but we did not investigated much about it). %Nevertheless, it is not surprising that \textit{normalization-equivariant} networks are more challenging to train as the proposed channel-wise sort pooling nonlinearities 
Furthermore, contrary to \cite{drunet} where the $\ell_1$ loss function is recommended to achieve better performance, supposedly due to its outlier robustness properties, we obtained slightly better results with the usual mean squared error (MSE) loss when dealing with \textit{normalization-equivariant} networks. Training was performed with a Quadro RTX 6000 GPU.

% RTX A5000

\begin{table*}[t]
\begin{center}
\caption{Execution time (in seconds) comparison on a training batch of size $16 \times 1 \times 128\times 128$ for different variants of the same DRUNet architecture \cite{drunet} (GPU: Quadro RTX 6000, CPU: 2,3 GHz Intel Core i7). The difference in speed between variants is more pronounced on GPU than on CPU.}

%\resizebox{1\columnwidth}{!}{%
\begin{tabular}{cc}
& \\
\multirow{4}{*}{\rotatebox[origin=c]{90}{\small \textbf{GPU}}}& \textit{(scale-equiv)} \\
&  \\
&  \\
& \textit{(norm-equiv)} \\
\end{tabular}
  \begin{NiceTabular}{ccll}

   Affine  & SortPool & \multicolumn{1}{c}{Backward pass $\downarrow$}  & \multicolumn{1}{c}{Inference pass $\downarrow$}\\[0.1cm]
   \hline\noalign{\vskip 0.1cm}

\textcolor{red}{\ding{55}} & \textcolor{red}{\ding{55}}   & \begin{tikzpicture}[scale=2]
\fill[opacity = 0.5, blue] (0.,0.05) -- (0.916, 0.05)   -- (0.916,-0.05)-- (0., -0.05) -- cycle; 
\end{tikzpicture} \small 0.229
& 
\begin{tikzpicture}[scale=2]
\fill[opacity = 0.5, red] (0.,0.05) -- (0.268, 0.05)   -- (0.268,-0.05)-- (0., -0.05) -- cycle; 
\end{tikzpicture} \small 0.067\\

%\textcolor{red}{\ding{55}} & \textcolor{red}{\ding{55}} & \textcolor{greenG}{\ding{51}}  & \begin{tikzpicture}[scale=2]\fill[opacity = 0.5, blue] (0.,0.05) -- (1.208, 0.05)   -- (1.208,-0.05)-- (0., -0.05) -- cycle; \end{tikzpicture} \small 0.302 & \begin{tikzpicture}[scale=2]\fill[opacity = 0.5, red] (0.,0.05) -- (0.308, 0.05)   -- (0.308,-0.05)-- (0., -0.05) -- cycle; \end{tikzpicture} \small 0.077\\

\textcolor{red}{\ding{55}} & \textcolor{greenG}{\ding{51}}  & \begin{tikzpicture}[scale=2]
\fill[opacity = 0.5, blue] (0.,0.05) -- (1.072, 0.05)   -- (1.072,-0.05)-- (0., -0.05) -- cycle; 
\end{tikzpicture} \small 0.268 & \begin{tikzpicture}[scale=2]
\fill[opacity = 0.5, red] (0.,0.05) -- (0.408, 0.05)   -- (0.408,-0.05)-- (0., -0.05) -- cycle; 
\end{tikzpicture} \small 0.102 \\

%\textcolor{red}{\ding{55}} & \textcolor{greenG}{\ding{51}} & \textcolor{greenG}{\ding{51}}  & \begin{tikzpicture}[scale=2]\fill[opacity = 0.5, blue] (0.,0.05) -- (1.372, 0.05)   -- (1.372,-0.05)-- (0., -0.05) -- cycle; \end{tikzpicture} \small 0.343 & \begin{tikzpicture}[scale=2]\fill[opacity = 0.5, red] (0.,0.05) -- (0.448, 0.05)   -- (0.448,-0.05)-- (0., -0.05) -- cycle; \end{tikzpicture} \small 0.112\\

%\textcolor{greenG}{\ding{51}} & \textcolor{red}{\ding{55}} & \textcolor{red}{\ding{55}}  & \begin{tikzpicture}[scale=2]\fill[opacity = 0.5, blue] (0.,0.05) -- (1.368, 0.05)   -- (1.368,-0.05)-- (0., -0.05) -- cycle; \end{tikzpicture} \small 0.342 & \begin{tikzpicture}[scale=2]\fill[opacity = 0.5, red] (0.,0.05) -- (0.292, 0.05)   -- (0.292,-0.05)-- (0., -0.05) -- cycle; \end{tikzpicture} \small 0.073  \\

\textcolor{greenG}{\ding{51}} & \textcolor{red}{\ding{55}}   & \begin{tikzpicture}[scale=2]
\fill[opacity = 0.5, blue] (0.,0.05) -- (1.376, 0.05)   -- (1.376,-0.05)-- (0., -0.05) -- cycle; 
\end{tikzpicture} \small 0.344 & \begin{tikzpicture}[scale=2]
\fill[opacity = 0.5, red] (0.,0.05) -- (0.332, 0.05)   -- (0.332,-0.05)-- (0., -0.05) -- cycle; 
\end{tikzpicture} \small 0.083 \\

%\textcolor{greenG}{\ding{51}} & \textcolor{greenG}{\ding{51}} & \textcolor{red}{\ding{55}}  & \begin{tikzpicture}[scale=2]\fill[opacity = 0.5, blue] (0.,0.05) -- (1.532, 0.05)   -- (1.532,-0.05)-- (0., -0.05) -- cycle; \end{tikzpicture} \small 0.383 & \begin{tikzpicture}[scale=2]\fill[opacity = 0.5, red] (0.,0.05) -- (0.448, 0.05)   -- (0.448,-0.05)-- (0., -0.05) -- cycle; \end{tikzpicture} \small 0.112 \\

\textcolor{greenG}{\ding{51}} & \textcolor{greenG}{\ding{51}}   & \begin{tikzpicture}[scale=2] \fill[opacity = 0.5, blue] (0.,0.05) -- (1.544, 0.05)   -- (1.544,-0.05)-- (0., -0.05) -- cycle; \end{tikzpicture} \small 0.386 & \begin{tikzpicture}[scale=2] \fill[opacity = 0.5, red] (0.,0.05) -- (0.488, 0.05)   -- (0.488,-0.05)-- (0., -0.05) -- cycle; \end{tikzpicture} \small 0.122\\[0.1cm]
   \hline\noalign{\vskip 0.1cm}
\end{NiceTabular}%
% }
\end{center}

\begin{center}
\begin{tabular}{c}
     \rotatebox[origin=c]{90}{\small \textbf{CPU}}
\end{tabular}
  \begin{NiceTabular}{cll}
Variant & \multicolumn{1}{c}{Backward pass $\downarrow$}  & \multicolumn{1}{c}{Inference pass $\downarrow$}\\[0.1cm]
   \hline\noalign{\vskip 0.1cm}

\textit{ordinary}   & \begin{tikzpicture}[scale=2]
\fill[opacity = 0.5, blue] (0.,0.05) -- (1.038, 0.05)   -- (1.038,-0.05)-- (0., -0.05) -- cycle; 
\end{tikzpicture} \small 38.4
& 
\begin{tikzpicture}[scale=2]
\fill[opacity = 0.5, red] (0.,0.05) -- (0.3389, 0.05)   -- (0.3389,-0.05)-- (0., -0.05) -- cycle; 
\end{tikzpicture} \small  12.5 \\

\textit{scale-equiv}  & \begin{tikzpicture}[scale=2]
\fill[opacity = 0.5, blue] (0.,0.05) -- (1.01, 0.05)   -- (1.01,-0.05)-- (0., -0.05) -- cycle; 
\end{tikzpicture} \small 37.4 & \begin{tikzpicture}[scale=2]
\fill[opacity = 0.5, red] (0.,0.05) -- (0.3295, 0.05)   -- (0.3295,-0.05)-- (0., -0.05) -- cycle; 
\end{tikzpicture} \small 12.2  \\

\textit{norm-equiv}   & \begin{tikzpicture}[scale=2]
\fill[opacity = 0.5, blue] (0.,0.05) -- (1.223, 0.05)   -- (1.223,-0.05)-- (0., -0.05) -- cycle; 
\end{tikzpicture} \small 45.3  & \begin{tikzpicture}[scale=2]
\fill[opacity = 0.5, red] (0.,0.05) -- (0.385, 0.05)   -- (0.385,-0.05)-- (0., -0.05) -- cycle; 
\end{tikzpicture} \small 14.3  \\[0.1cm]
   \hline\noalign{\vskip 0.1cm}
\end{NiceTabular}%
% }
\end{center}

\small Affine: affine-constrained convolutions with reflect padding and affine residual connections.

\small SortPool: channel-wise sort pooling nonlinearities instead of ReLU.

\label{timeComp}
\end{table*}

\textcolor{black}{In Table \ref{timeComp}, we compare the computational costs of different variants for training and inference. Interestingly, the computational cost on GPU for training is much more sensitive to the ``affine mode'' (involving affine-constrained convolutions with reflect padding and affine residual connection) than to sort pooling nonlinearities, while it is the opposite for inference. All in all, for gaining \textit{normalization-equivariance}, the learning and inference time is almost doubled for the DRUNet architecture \cite{drunet} on GPU. Note however that we do not claim to have the most optimized implementation and there is probably room for improvement. Surprisingly, the difference in speed between all variants is much less pronounced on CPU. In particular, the inference pass takes only about  $20\%$ longer for the \textit{normalization-equivariant} DRUNet on CPU.
}

\section{Mathematical proofs for normalization-equivariant neural networks}

\subsection{Proofs of Lemmas and Propositions}
\label{proofs_charac}

\noindent \textbf{Lemma \ref{characterization}}(Characterizations)
\begin{proof} For each type of equivariance, both existence and uniqueness of $f$ must be proven.
Let $\mathbf{0}_n$ be the zero vector of $\mathbb{R}^n$ and $(y_x)_{x\in \mathcal{C}}$ the values that $f$ takes on its characteristic set $\mathcal{C}$. 

\textit{Scale-equivariance:} 

\begin{itemize}
\item Uniqueness: Let $f$ and $g$ two \textit{scale-equivariant} functions such that $\forall x \in \mathcal{S}, f(x) = g(x)$.  First of all, for any  \textit{scale-equivariant} function $h$, $h(\mathbf{0}_n) = h(2 \cdot \mathbf{0}_n) = 2 h(\mathbf{0}_n)$, hence $h(\mathbf{0}_n) = \mathbf{0}_m$. Therefore, $f(\mathbf{0}_n) = g(\mathbf{0}_n) = \mathbf{0}_m$.

Let $x \in  \mathbb{R}^n \setminus \{ \mathbf{0}_n \}$. As $ \frac{x}{\|x\|} \in \mathcal{S}$,  we have $ f(\frac{x}{\|x\|}) = g(\frac{x}{\|x\|}) \Rightarrow \frac{1}{\|x\|} f(x) = \frac{1}{\|x\|} g(x) \Rightarrow f(x) =g(x)$. Finally, $f=g$.
\item Existence: Let $f : x \in \mathbb{R}^n \mapsto \left\{
    \begin{array}{ll}
        \|x\| \cdot y_{\frac{x}{\|x\|}}  & \mbox{if } x \neq \mathbf{0}_n \\
        \mathbf{0}_m & \mbox{otherwise}
    \end{array}
\right.\,.$ Note that $\forall x \in \mathcal{S}, f(x) = y_x$.  Let $x \in \mathbb{R}^n$ and $ \lambda \in \mathbb{R}^+_\ast$. If $x \neq \mathbf{0}_n$, $f(\lambda x) = \|\lambda x\| \cdot y_{\frac{\lambda x}{\|\lambda x\|}} =  \lambda \| x\| \cdot y_{\frac{ x}{\| x\|}} = \lambda f(x)$ and if $x = \mathbf{0}_n$ $f(\lambda x) = \mathbf{0}_m = \lambda f(x)$,  hence $f$ is \textit{scale-equivariant}.
\end{itemize}

\textit{Shift-equivariance:}

\begin{itemize}
\item Uniqueness:  Let $f$ and $g$ two \textit{shift-equivariant} functions such that $\forall x \in \operatorname{Span}(\mathbf{1}_n)^\bot, f(x) = g(x)$. Let $x \in \mathbb{R}^n$. By orthogonal decomposition of $\mathbb{R}^n$ into  $\operatorname{Span}(\mathbf{1}_n)^\bot$ and $\operatorname{Span}(\mathbf{1}_n)$: $$\exists! \, (x_1, x_2) \in  \operatorname{Span}(\mathbf{1}_n)^\bot \times \operatorname{Span}(\mathbf{1}_n), \: x = x_1 + x_2\,.$$
Then, $f(x) = f(x_1 + x_2) = f(x_1) + x_2 = g(x_1) + x_2 = g(x_1 + x_2) = g(x)$.
\item Existence: Let $f : x \in \mathbb{R}^n \mapsto y_{x_1} + x_2$, where $x = x_1 + x_2$ is the unique decomposition such that $x_1 \in \operatorname{Span}(\mathbf{1}_n)^\bot$ and $x_2 \in \operatorname{Span}(\mathbf{1}_n)$. Note that $\forall x \in \operatorname{Span}(\mathbf{1}_n)^\bot, f(x) = y_x$. Let $x \in \mathbb{R}^n$ and $\mu \in \mathbb{R}$. $f(x + \mu) = y_{x_1} + x_2 + \mu \mathbf{1}_m = f(x)+ \mu$ as if $x$ orthogonally decomposes into $x_1 + x_2$ with $x_1 \in \operatorname{Span}(\mathbf{1}_n)^\bot$ and $x_2 \in \operatorname{Span}(\mathbf{1}_n)$, then $x + \mu$ orthogonally decomposes into $x_1 + (x_2 + \mu \mathbf{1}_m)$. $f$ is then $\textit{shift-equivariant}$.
\end{itemize}

\textit{Normalization-equivariance:}

\begin{itemize}
\item Uniqueness: Let $f$ and $g$ two \textit{normalization-equivariant} functions such that $\forall x \in \mathcal{S} \cap \operatorname{Span}(\mathbf{1}_n)^\bot, f(x) = g(x)$. 
First, as $f$ and $g$ are \textit{a fortiori} \textit{scale-equivariant}, $f(\mathbf{0}_n) = g(\mathbf{0}_n) = \mathbf{0}_m$.
Let $x \in \mathbb{R}^n \setminus \{ \mathbf{0}_n \}$. By orthogonal decomposition of $\mathbb{R}^n$ into  $\operatorname{Span}(\mathbf{1}_n)^\bot$ and $\operatorname{Span}(\mathbf{1}_n)$: $$\exists! \, (x_1, x_2) \in  \operatorname{Span}(\mathbf{1}_n)^\bot \times \operatorname{Span}(\mathbf{1}_n), \: x = x_1 + x_2\,.$$

If $x_1 = \mathbf{0}_n$, $f(x) = f(\mathbf{0}_n + x_2)= f(\mathbf{0}_n) + x_2 = \mathbf{0}_m + x_2 = x_2$. Likewise, $g(x) = x_2$, hence $f(x) = g(x)$. Else, if $x_1 \neq \mathbf{0}_n, f(x) = f(x_1 + x_2) = f(x_1) + x_2  = \|x_1\| f(\frac{x_1}{\|x_1\|}) + x_2 = \|x_1\| g(\frac{x_1}{\|x_1\|}) + x_2 = g(x_1) + x_2 = g(x_1 + x_2) = g(x)$, as $\frac{x_1}{\|x_1\|} \in \mathcal{S} \cap \operatorname{Span}(\mathbf{1}_n)^\bot$. Finally, $f=g$.

\item Existence: Let $f : x \in \mathbb{R}^n \mapsto \left\{
    \begin{array}{ll}
        \|x_1\| \cdot y_{\frac{x_1}{\|x_1\|}} + x_2  & \mbox{if } x_1 \neq \mathbf{0}_n \\
        x_2 & \mbox{otherwise}
    \end{array}
\right.$, where $x = x_1 + x_2$ is the unique decomposition such that $x_1 \in \operatorname{Span}(\mathbf{1}_n)^\bot$ and $x_2 \in \operatorname{Span}(\mathbf{1}_n)$.  Note that $\forall x \in \mathcal{S} \cap \operatorname{Span}(\mathbf{1}_n)^\bot, f(x) = y_x$. Let $x \in \mathbb{R}^n$, $\lambda \in \mathbb{R}^+_\ast$ and $\mu \in \mathbb{R}$. $x$ decomposes orthogonally into $x_1 + x_2$ with $x_1 \in \operatorname{Span}(\mathbf{1}_n)^\bot$ and $x_2 \in \operatorname{Span}(\mathbf{1}_n)$, and we have $f(\lambda x + \mu) = f(\lambda x_1 + (\lambda x_2  + \mu))$, where $\lambda x_1 + (\lambda x_2  + \mu)$ is the orthogonal decomposition of $\lambda x + \mu$ into  $\operatorname{Span}(\mathbf{1}_n)^\bot$ and $\operatorname{Span}(\mathbf{1}_n)$.

If $x_1 = \mathbf{0}_n $, then $\lambda x_1 = \mathbf{0}_n$ and $f(\lambda x + \mu) = \lambda x_2  + \mu = \lambda f(x)  + \mu$.

Else, if $x_1 \neq \mathbf{0}_n$, then $\lambda x_1 \neq \mathbf{0}_n$, and $f(\lambda x + \mu) =  \|\lambda x_1\| \cdot y_{\frac{\lambda x_1}{\|\lambda x_1\|}}  + (\lambda x_2  + \mu) = \lambda (\| x_1\| \cdot y_{\frac{x_1}{\|x_1\|}}  +  x_2)  + \mu = \lambda f(x) + \mu $. Finally, $f$ is \textit{normalization-equivariant}.
\end{itemize}

\end{proof}

\noindent \textbf{Lemma \ref{preserving}} (Operations preserving equivariance)

\begin{proof}
Let $x \in \mathbb{R}^n$, $\lambda \in \mathbb{R}^+_\ast$ and $\mu \in \mathbb{R}$.

\begin{itemize}
\setlength\itemsep{-0.4em}
\item If $f$ and $g$ are both \textit{scale-equivariant}, $(f \circ g)(\lambda x) = f(g(\lambda x)) = f(\lambda g(x)) = \lambda f( g(x)) = \lambda (f \circ g)(x)$ and if they are both \textit{shift-equivariant}, $(f \circ g)(x + \mu) = f(g(x + \mu)) = f(g(x) + \mu) =  f( g(x)) + \mu = (f \circ g)(x) + \mu$.
\item Let $h : x \mapsto ( f(x)^\top  \: g(x)^\top )^\top$. If $f$ and $g$ are both \textit{scale-equivariant}, $h(\lambda x) = ( f(\lambda x)^\top  \: g(\lambda x)^\top )^\top = ( \lambda f( x)^\top  \: \lambda g( x)^\top )^\top = \lambda h(x)$ and if they are both \textit{shift-equivariant}, $h(x + \mu) = ( f(x+\mu)^\top  \: g(x+\mu)^\top )^\top = ( f(x)^\top +\mu  \quad g(x)^\top  +\mu )^\top  = h(x) + \mu$.
\item Let $t \in \mathbb{R}$ and $h: x \mapsto (1-t)  f +  t g$. If $f$ and $g$ are both \textit{scale-equivariant}, $h(\lambda x) = (1-t)  f(\lambda x) +  t g(\lambda x) = (1-t)  \lambda f( x) +  t \lambda g( x) = \lambda ((1-t)  f( x) +  t g( x)) = \lambda h(x)$ and if they are both \textit{shift-equivariant}, $h(x+\mu) = (1-t)  f(x+\mu) +  t g(x+\mu) = (1-t)  (f( x) + \mu) +  t  (g( x) + \mu) = (1-t)  f( x) +  t g( x) + (1-t)  \mu +  t \mu =  h(x) + \mu$.
\end{itemize} 
\end{proof}

\noindent \textbf{Proposition \ref{case_of_nn}}

\begin{proof}
    Let $a < b \in \mathbb{R}$, $f :  \mathbb{R}^n \mapsto \mathbb{R}^m$, $x \in \mathbb{R}^n$, $\lambda \in \mathbb{R}^+_\ast$ and $\mu \in \mathbb{R}$. %Let $x_{min} = \min( x )$ and $x_{max} = \max(x)$. 
    
    We have $\mathcal{T}_{a , b}(\lambda x + \mu) = (b-a) \frac{\lambda x + \mu - \min(\lambda x + \mu)}{\max(\lambda x + \mu) - \min(\lambda x + \mu)} + a = (b-a) \frac{x - \min(x)}{\max(x) - \min(x)} + a = \mathcal{T}_{a , b}(x)$ (\textit{i.e.} $\mathcal{T}_{a , b}$ is \textit{normalization-invariant}).
    $\mathcal{T}^{-1}_{a , b}$ denotes the inverse transformation intricately linked to the input $x$ of $\mathcal{T}_{a , b}$ (note that this is an improper notation as $\mathcal{T}_{a , b}$ is not bijective). Thus, if $x$ is the input of $\mathcal{T}_{a , b}$, then $\mathcal{T}^{-1}_{a , b} : y \mapsto (\max(x) - \min(x)) \frac{y - a}{b-a} + \min(x)$.
    
    %and $\mathcal{T}^{-1}_{a , b} : x  \mapsto (x_{max} - x_{min}) \frac{x - a}{b-a} + x_{min}$.
    
\begingroup \allowdisplaybreaks \begin{align*}
        (\mathcal{T}^{-1}_{a , b} \circ f \circ \mathcal{T}_{a , b})(\lambda x + \mu) &=  \frac{\max(\lambda x + \mu) - \min(\lambda x + \mu)}{b-a} \left( (f \circ \mathcal{T}_{a , b})(\lambda x + \mu) - a \right)+ \min(\lambda x + \mu) \,, \\
        &= \lambda \frac{\max( x ) - \min(x)}{b-a} \left( (f \circ \mathcal{T}_{a , b})(\lambda x + \mu) - a \right)+ \lambda \min(x ) + \mu \,, \\
        & =\lambda \left(\frac{\max( x ) - \min(x)}{b-a} \left( (f \circ \mathcal{T}_{a , b})(x) - a \right)+  \min(x )\right) + \mu  \,, \\
        &= \lambda (\mathcal{T}^{-1}_{a , b} \circ f \circ \mathcal{T}_{a , b})(x) + \mu\,.
\end{align*} \endgroup
  \noindent  Finally, $\mathcal{T}^{-1}_{a , b} \circ f \circ \mathcal{T}_{a , b}$ is \textit{normalization-equivariant}. \end{proof}

\color{black}
\noindent \textbf{Proposition \ref{set_prop}}

\begin{proof} 

Let $\operatorname{id} : x \in \mathbb{R} \mapsto x$ be the identity function. $\operatorname{id}$ is a \textit{normalization-equivariant} function so  $\{ x \mapsto x \} \subseteq\operatorname{NE}(1)$. Reciprocally, let $f \in \operatorname{NE}(1)$. By \textit{scale-equivariance}, $f(0) = f(2 \times 0) = 2 f(0)$, hence $f(0) = 0$. By \textit{shift-equivariance}, $\forall x \in \mathbb{R}, f(x) = f(x + 0) = f(0) + x = x$, hence,  $f=\operatorname{id}$. Finally, $\operatorname{NE}(1)  \subseteq \{ x \mapsto x \}$, hence $\operatorname{NE}(1)  = \{ x \mapsto x \}$. Note that it is coherent with Lemma \ref{characterization} which states that $f$ is entirely determined by its values on $\mathcal{S} \cap \operatorname{Span}(\mathbf{1}_n)^\bot$, which reduces to the empty set for $n=1$. 

Let $F = \left \{  {\left. (x_1, x_2)  \mapsto A \begin{pmatrix} x_1 \\ x_2 \end{pmatrix}  \mbox{ if } x_1 \leq x_2 \mbox{ else }  B \begin{pmatrix}    x_1 \\ x_2 \end{pmatrix}     
 \,\right|\, A, B \in \mathbb{R}^{2 \times 2}} \;\text{s.t.}\; A \mathbf{1}_2 = B \mathbf{1}_2 = \mathbf{1}_2 \right \}$ and $f \in F$. Let $x \in \mathbb{R}^2$, $\lambda > 0$ and $\mu \in \mathbb{R}$. 
 
 $f(\lambda x + \mu) = \left\{
    \begin{array}{ll}
        A (\lambda x + \mu) & \mbox{if } \lambda x_1 + \mu \leq  \lambda x_2 + \mu \\
        B (\lambda x + \mu) & \mbox{otherwise}
    \end{array} \right. =  \left\{ \begin{array}{ll}
       \lambda A  x  + \mu  & \mbox{if }  x_1  \leq   x_2 \\
        \lambda B  x  + \mu  & \mbox{otherwise}
    \end{array} \right. = \lambda f(x) + \mu,$ hence $f \in \operatorname{NE}(2)$. 
    
    Reciprocally, let $f \in \operatorname{NE}(2)$. For $n=2$ and when considering the Euclidean distance, $\mathcal{S} \cap \operatorname{Span}(\mathbf{1}_n)^\bot = \{-u, u\}$ with $u = (-1 / \sqrt{2}, 1 / \sqrt{2})$. Let
    
     $\displaystyle A = \frac{1}{u_2 - u_1} \begin{pmatrix}
        u_2 - f(u)_1 & f(u)_1 - u_1 \\
        u_2 - f(u)_2 & f(u)_2 - u_1
    \end{pmatrix}$ and $\displaystyle B = \frac{1}{u_2 - u_1} \begin{pmatrix}
        u_2 + f(-u)_1 & -f(-u)_1 - u_1 \\
        u_2 + f(-u)_2 & -f(-u)_2 - u_1
    \end{pmatrix}$.
    Let $g : (x_1, x_2)  \mapsto A \begin{pmatrix} x_1 \\ x_2 \end{pmatrix}  \mbox{ if } x_1 \leq x_2 \mbox{ else }  B \begin{pmatrix}    x_1 \\ x_2 \end{pmatrix}$. We have $g \in F$ since $A \mathbf{1}_2 = B \mathbf{1}_2 = \mathbf{1}_2$ (in particular $g$ is then \textit{normalization-equivariant}) and $\forall x \in \{-u, u\}, g(x) = f(x)$. According to Lemma \ref{characterization}, $g=f$, hence $f \in F$. Finally,  $\operatorname{NE}(2) \subseteq F$, hence $\operatorname{NE}(2) = F$. 
\end{proof}

\color{black}

\noindent \textbf{Proposition \ref{theorem}}

\begin{proof}
    $f_\theta^{\text{NE}}$ is composed of \textcolor{black}{three} types of building blocks of the following form:
    
\begin{itemize}
    \item affine convolutions: $a_\Theta : x \in \mathbb{R}^n \mapsto \Theta x$ \: with $\Theta \in \mathbb{R}^{m \times n}$ subject to $\Theta \mathbf{1}_n = \mathbf{1}_m$\,,
    \item sort pooling nonlinearities: $\operatorname{sortpool} :  \mathbb{R}^n \mapsto \mathbb{R}^n$\,,
    \item \textcolor{black}{max pooling layers: $\operatorname{maxpool} :  \mathbb{R}^n \mapsto \mathbb{R}^m$ with $m< n$    \,,}
\end{itemize}

which are assembled using:

\begin{itemize}
    \item function compositions: $\operatorname{comp}(f, g) \mapsto f \circ g$\,,
    \item skip connections: $\operatorname{skip}(f, g) \mapsto (x \mapsto ( f(x)^\top  \: g(x)^\top )^\top)$\,,
    \item affine residual connections: $\operatorname{ares}_t(f, g) \mapsto (1-t) f + tg$ with $t \in \mathbb{R}$\,.
\end{itemize}

Note that the rows of $\Theta$ in $a_\Theta$ encode the convolution kernels in a CNN and the trainable parameters, denoted by $\theta$, are only composed of matrices $\Theta$ and scalars $t$.  \textcolor{black}{Moreover, note that average pooling layers are nothing else than affine convolutions with fixed parameters}.

Since $a_\Theta$, $\operatorname{sortpool}$ and \textcolor{black}{$\operatorname{maxpool}$} are \textit{normalization-equivariant} functions, \textcolor{black}{Lemma} \ref{preserving} states that the resulting function $f_\theta^{\text{NE}}$ is also \underline{\textit{normalization-equivariant}}. Moreover, since they are continuous and the assembling operators preserve continuity, $f_\theta^{\text{NE}}$ is \underline{continuous}. Then, for a given input $x \in \mathbb{R}^n$, we have $(\operatorname{sortpool} \; \circ \; a_\Theta)(x) = a_{\pi(\Theta)}(x) = \pi(\Theta)x$, where $\pi$ an operator acting on matrix $\Theta$ by permuting its rows (note that the permutation $\pi$ is both dependent on $x$ and $\Theta$). Therefore, applying a pattern ``conv affine + sortpool'' simply amounts locally to a linear transformation. \textcolor{black}{Moreover, since applying a max pooling layer amounts to removing some rows from matrix $\Theta$, the local linear behavior is preserved}. Thus, as the nonlinearities of $f_\theta^{\text{NE}}$ are exclusively brought by sort pooling patterns \textcolor{black}{(and possibly max pooling layers)}, $f_\theta^{\text{NE}}$ is actually locally linear. In other words,  $f_\theta^{\text{NE}}$ is \underline{piecewise-linear}. Moreover, as there is a finite number (although high) of possible permutations \textcolor{black}{(and possibly eliminations)} of the rows of all matrices $\Theta$, $f_\theta^{\text{NE}}$  has \underline{finitely many pieces}. Finally, on each piece represented by the vector $y_r$,  $f_\theta^{\text{NE}}(y) = A_\theta^{y_{r}} y$. It remains to prove that $A_\theta^{y_{r}} \mathbf{1}_n = \mathbf{1}_m$. But this property is easily obtained by noticing that, subject to dimensional compatibility on matrices $\Theta$:

\begin{itemize}
    \item $\Theta \mathbf{1}_n = \mathbf{1}_m \Rightarrow \pi(\Theta) \mathbf{1}_n = \mathbf{1}_m$ (``conv affine + sortpool'')\,,
    \textcolor{black}{\item $\Theta \mathbf{1}_{n} = \mathbf{1}_{m} \Rightarrow \rho(\Theta) \mathbf{1}_n = \mathbf{1}_l$ (``conv affine + maxpool'') where $\rho$ removes some rows\,,}
    \item $\Theta_1 \mathbf{1}_n = \mathbf{1}_m \text{ and } \Theta_2 \mathbf{1}_m = \mathbf{1}_l \Rightarrow \Theta_2 \Theta_1 \mathbf{1}_n = \mathbf{1}_l$ (composition)\,,
    \item $\Theta_1 \mathbf{1}_{n_1} = \mathbf{1}_{m_1} \text{ and } \Theta_2 \mathbf{1}_{n_2} = \mathbf{1}_{m_2} \Rightarrow \begin{pmatrix}
        \Theta_1 \\ \Theta_2
    \end{pmatrix}  \mathbf{1}_{n_1 + n_2} = \mathbf{1}_{m_1 + m_2}$ (skip connection)\,,
    \item $\Theta_1 \mathbf{1}_n = \mathbf{1}_m \text{ and } \Theta_2 \mathbf{1}_n = \mathbf{1}_m \Rightarrow (1-t) \Theta_1 \mathbf{1}_n + t \Theta_2 \mathbf{1}_n = \mathbf{1}_m$ (affine residual connection)\,.
\end{itemize}

Thus,  the affine combinations are preserved all along the layers of $f_\theta^{\text{NE}}$. In the end, 
    \begin{center}
        $f_\theta^{\text{NE}}(y) = A_\theta^{y_{r}} y, \: \text{ with } A_\theta^{y_{r}} \in \mathbb{R}^{m \times n} \text{ such that } A_\theta^{y_{r}} \mathbf{1}_n = \mathbf{1}_m\,.$
    \end{center}  \end{proof}

\subsection{Examples of normalization-equivariant conventional denoisers}
\label{proofs_ne}

\paragraph{Noise-reduction filters:}

All linear smoothing filters can be put under the form $f_\Theta : x \in \mathbb{R}^n \mapsto \Theta x$ with $\Theta \in \mathbb{R}^{n \times n}$ (the rows of $\Theta$ encode the convolution kernel). Obviously, $f_\Theta$ is always \textit{scale-equivariant}, whatever the filter $\Theta$. As for the \textit{shift-equivariance}, a simple calculation shows that:
$$ x\mapsto \Theta x \text{ is } \textit{shift-equivariant} \: \Leftrightarrow \: \forall x \in \mathbb{R}^n, \forall \mu \in \mathbb{R}, \Theta(x + \mu  \mathbf{1}_n) = \Theta x + \mu \mathbf{1}_m \: \Leftrightarrow \: \Theta \mathbf{1}_n = \mathbf{1}_m\,.$$
Since the sum of the coefficients of a Gaussian kernel and an averaging kernel is one, we have $\Theta \mathbf{1}_n = \mathbf{1}_m$, hence these linear filters are  \textit{normalization-equivariant}. 
The median filter is also 
\textit{normalization-equivariant} because $\operatorname{median}(\lambda x + \mu) = \lambda \operatorname{median}( x ) + \mu$ for $\lambda \in \mathbb{R}^+_\ast$ and $\mu \in \mathbb{R}$.

\paragraph{Patch-based denoising:} \textcolor{white}{-}

$-$ NLM \cite{nlmeans}: Assuming that the smoothing parameter $h$ is proportional to $\sigma$, \textit{i.e.} $h = \alpha \sigma$, we have
$e^{-\frac{\|p(\lambda y_i + \mu) - p(\lambda y_j + \mu)\|_2^2}{(\alpha \lambda \sigma)^2}} = e^{-\frac{\lambda^2 \|p(y_i) - p(y_j)\|_2^2}{\lambda^2 (\alpha \sigma)^2}} = e^{-\frac{\|p(y_i) - p(y_j)\|_2^2}{h^2}}$, hence the aggregation weights are \textit{normalization-invariant}. Then,
\begingroup \allowdisplaybreaks \begin{align*}
    f_{\operatorname{NLM}}(\lambda y + \mu, \lambda \sigma)_i &= \frac{1}{W_i}\sum_{y_j \in \Omega(y_i)} e^{-\frac{\|p(y_i) - p(y_j)\|_2^2}{h^2}} (\lambda y_j + \mu) \: \text{with} \,  W_i = \sum_{y_j \in \Omega(y_i)} e^{-\frac{\|p(y_i) - p(y_j)\|_2^2}{h^2}} \,, \\
    &= \frac{1}{W_i}  \sum_{y_j \in \Omega(y_i)} e^{-\frac{\|p(y_i) - p(y_j)\|_2^2}{h^2}} \lambda y_j + \frac{1}{W_i}  \sum_{y_j \in \Omega(y_i)} e^{-\frac{\|p(y_i) - p(y_j)\|_2^2}{h^2}}  \mu \,, \\
    &= \lambda \left( \frac{1}{W_i} \sum_{y_j \in \Omega(y_i)}  e^{-\frac{\|p(y_i) - p(y_j)\|_2^2}{h^2}}  y_j \right)  + \mu  \,, \\
    &= \lambda  f_{\operatorname{NLM}}(y, \sigma)_i  + \mu \,.
 \end{align*} \endgroup
Finally, $f_{\operatorname{NLM}}$ is a \textit{normalization-equivariant} function.

$-$ NL-Ridge \cite{nlridge}: The block-matching procedure at the heart of NL-Ridge is \textit{normalization-invariant} as it is based on comparisons of the $\ell_2$ norm of the difference of image patches. For each noisy patch group, \textit{a.k.a.} similarity matrix, $Y  \in \mathbb{R}^{n \times k}$ composed of $k$ vectorized similar patches of size $n$, the optimal weights $\Theta^\ast \in \mathbb{R}^{k \times k}$, in the $\ell_2$ risk sense, are computed such that $Y\Theta^\ast$ is as close as possible to the (unknown) clean patch group $X \in \mathbb{R}^{n \times k}$. The two successive minimization problems approximating $\Theta$ under affine constraints $\mathcal{C} = \{ \Theta \in \mathbb{R}^{k \times k},  \Theta^\top \mathbf{1}_k = \mathbf{1}_k\}$ can be put under the form:
$$
 \Theta^\ast = \arg \min_{\Theta \in \mathcal{C}} \; \operatorname{tr}\left( \frac{1}{2}  \Theta^\top Q \Theta  + C \Theta \right) = I_k - n\sigma^2 \left[Q^{-1} - \frac{Q^{-1} \mathbf{1}_k (Q^{-1} \mathbf{1}_k^\top)}{\mathbf{1}_k^\top Q^{-1} \mathbf{1}_k}   \right] \,. 
$$
with $Q = Y^\top Y$ or $Q = \hat{X}^\top \hat{X} + n\sigma^2 I_k$ for the first and second step, respectively ($\hat{X}$ is the patch group estimate obtained after the first step), $C = n\sigma^2 I_k - Q$ and where $\operatorname{tr}$ denotes the trace operator. Depending on the step, we have:

$$
2 \operatorname{tr}\left( \frac{1}{2}  \Theta^\top Q \Theta  + C \Theta \right)=  \left\{
    \begin{array}{l}
    \| Y\Theta - Y \|^2_F  + 2 n\sigma^2 \operatorname{tr}\left(\Theta\right) + \operatorname{const}
    \\\text{or}\\
         \| \hat{X}\Theta - \hat{X} \|^2_F  + n\sigma^2 \| \Theta \|_F^2  + \operatorname{const} 
         
    \end{array}
\right.
$$
where $\| . \|_F$ is the Frobenius norm. But, for any $Z \in \mathbb{R}^{n \times k}$ and any function $h : \Theta \in \mathbb{R}^{k \times k} \mapsto \mathbb{R}$,  $$\| (\lambda Z + \mu) \Theta - (\lambda Z + \mu) \|^2_F  + n(\lambda \sigma)^2 h(\Theta) = \lambda^2 \left( \|  Z  \Theta -  Z  \|^2_F  + n\sigma^2 h(\Theta) \right)\,,$$
assuming that $\Theta^\top \mathbf{1}_k = \mathbf{1}_k$. Therefore, the aggregation weights $\Theta^\ast$ are \textit{normalization-invariant} and $(\lambda Y + \mu) \Theta^\ast = \lambda Y\Theta^\ast + \mu$. Finally, NL-Ridge with affine constraints encodes a \textit{normalization-equivariant} function.

\paragraph{TV denoising:}  %By definition, $f_{\operatorname{TV}}(y, \sigma) = \mathop{\arg \min}\limits_{\substack{x \in \mathbb{R}^{n}}} \: \| x \|_{\operatorname{TV}}  \quad \text{s.t.} \quad  \| y-x \|_2^2  = n\sigma^2\,.$ 

Let $y \in \mathbb{R}^n, \lambda \in \mathbb{R}^+_\ast$ and $\mu \in \mathbb{R}$. Let $x^\ast = \mathop{\arg \min}\limits_{
\substack{x \in \mathbb{R}^{n}}} \: \| x \|_{\operatorname{TV}}  \quad \text{s.t.} \quad  \| y-x \|_2^2  = n\sigma^2$ be the solution of TV \cite{TV}.
\begingroup \allowdisplaybreaks \begin{align*}
    f_{\operatorname{TV}}(\lambda y + \mu, \lambda \sigma) &= \mathop{\arg \min}\limits_{
\substack{x \in \mathbb{R}^{n}}} \: \| x \|_{\operatorname{TV}}  \quad \text{s.t.} \quad  \| \lambda y + \mu -x \|_2^2  = n(\lambda\sigma)^2 \,, \\
&= \mathop{\arg \min}\limits_{
\substack{x \in \mathbb{R}^{n}}} \: \lambda \left\lVert \frac{x - \mu}{\lambda} \right\rVert_{\operatorname{TV}}  \quad \text{s.t.} \quad  \lambda^2 \left\lVert  y - \frac{x - \mu}{\lambda} \right\rVert_2^2  = \lambda^2 n\sigma^2 \,, \\
&= \mathop{\arg \min}\limits_{
\substack{x \in \mathbb{R}^{n}}} \:  \left\lVert \frac{x - \mu}{\lambda} \right\rVert_{\operatorname{TV}}  \quad \text{s.t.} \quad   \left\lVert  y - \frac{x - \mu}{\lambda} \right\rVert_2^2  = n\sigma^2 \,, \\
&= \lambda x^\ast + \mu \,,\\
&= \lambda f_{\operatorname{TV}}(y, \sigma) + \mu \,.
  \end{align*} \endgroup
Finally, $f_{\operatorname{TV}}$ is a \textit{normalization-equivariant} function.

\section{Additional results}
\label{appendix_C}

\begin{table*}[h]
\centering
\caption{Real-world image denoising on Darmstadt Noise Dataset (DND) with raw data and variance-stabilizing transformation (VST). All ``non-blind'' methods were trained solely on synthetic white Gaussian noise.}

\resizebox{1\columnwidth}{!}{%
  \begin{NiceTabular}{c@{\hspace{0.1cm}}  c@{\hspace{0.5cm}}c@{\hspace{0.7cm}}c@{\hspace{0.0cm}} c@{\hspace{0.99cm}}          c@{\hspace{0.1cm}}  c@{\hspace{0.5cm}}c@{\hspace{0.7cm}}c@{\hspace{0.0cm}} c}

 \multicolumn{2}{c}{Quality metric} & PSNR $\uparrow$ &  SSIM $\uparrow$ & & \multicolumn{2}{c}{Quality metric} & PSNR $\uparrow$ &  SSIM $\uparrow$ & \\[0.1cm]
   \hline\noalign{\vskip 0.1cm}

 \multirow{3}{*}{\small DRUNet} \quad \quad  & \textit{ordinary}   &  47.58  &  0.9762  & &  \multirow{3}{*}{\small FDnCNN} \quad \quad  & \textit{ordinary}  &   47.37  &   0.9754  &\\
 
 & \textit{scale-equiv}   &   47.59 & 0.9763 & & & \textit{scale-equiv}   &   47.31 & 0.9754  &\\

 & \textbf{\textit{norm-equiv}}  & 47.57  & 0.9762   & & & \textbf{\textit{norm-equiv}}  &   47.35 & 0.9753 & \\[0.1cm] \hline
\end{NiceTabular}%

%\cdashline{1-4}\noalign{\vskip 0.1cm}
%\hline\noalign{\vskip 0.1cm}

}

\label{resultsPSNR2}
\end{table*}

\input{figure_generalization_FDnCNN}
\input{tableau_figure_generalization}
\input{figure_house_FDnCNN}
\input{figure_monarch_FDnCNN}
\input{figure_photo}

%Test error $\downarrow$

\begin{table*}[p]
\begin{center}
\caption{Test error of different variants of the same VGG8b architecture for image classification. Note that norm-equivariance is with respect to the classification vector and becomes norm-invariance after label selection via argmax. See \cite{nokland} for details about architecture, datasets and training.}

%\resizebox{1\columnwidth}{!}{%
  \begin{NiceTabular}{c@{\hspace{0.1cm}}  c@{\hspace{0.5cm}}c@{\hspace{0.7cm}}c@{\hspace{0.7cm}}c@{\hspace{0.7cm}}c@{\hspace{0.0cm}}}

 \multicolumn{2}{c}{Dataset} & MNIST    & Kuzushiji-MNIST  & Fashion-MNIST \\[0.1cm]
   \hline\noalign{\vskip 0.1cm}

 \multirow{3}{*}{\small VGG8b} \quad \quad  & \textit{ordinary}  & 0.26\%   & 1.53\% &  4.53\% &  \\
 
 & \textit{scale-equiv}  & 0.37\% & 1.52\% &   5.01\%  & \\
 
 & \textbf{\textit{norm-equiv}} &  0.44\%  & 2.78\% & 6.57\%  &   \\[0.1cm] \hline
\end{NiceTabular}%

\end{center}
%\cdashline{1-4}\noalign{\vskip 0.1cm}
%\hline\noalign{\vskip 0.1cm}

  \bigskip 
  \small Note that for \textbf{\textit{norm-equiv}} variants, learning rate is initialized to $3e\hbox{-}5$ instead of $5e\hbox{-}4$ and dropout rate is halved.

%}

\label{mnist}
\end{table*}

%\newpage
%\input{instructions}

\end{document}